\documentclass{article}

\usepackage{microtype}
\usepackage{graphicx}
\usepackage{subcaption}
\usepackage{booktabs} 

\usepackage{hyperref}

\usepackage{enumitem}

\usepackage{booktabs}
\usepackage{multicol}

\usepackage{tabularx}  
\usepackage{siunitx}
\usepackage{float}  
\usepackage{tikz}

\usepackage{amsmath, amssymb, bm}
\usepackage{graphicx}

\newcommand{\norm}[1]{\left\lVert #1 \right\rVert}

\newcommand{\psiin}{\psi_{\text{in}}}
\newcommand{\psiout}{\psi_{\text{out}}}

\newcommand{\token}[1]{\texttt{\small #1}}
\newcommand{\nnEmbedding}[0]{\hat{z}}
\newcommand{\maxDistance}[0]{\delta}
\newcommand{\fairnessScore}[0]{\Psi}
\usepackage[sets,operations,nn,colors]{cora}

\usepackage{makecell}
\usepackage{amsmath}
\usepackage{amssymb}
\usepackage[table,xcdraw]{xcolor}
\usepackage{cleveref}
\usepackage{tcolorbox}

\usepackage[preprint]{icml2026}

\usepackage{amsmath}
\usepackage{amssymb}
\usepackage{mathtools}
\usepackage{amsthm}
\usepackage{multirow}

\usepackage{algorithm}
\usepackage{algorithmic}
\usepackage{xurl}

\theoremstyle{plain}

\theoremstyle{definition}

\theoremstyle{remark}

\usepackage[textsize=tiny]{todonotes}

\icmltitlerunning{Language Models That Walk the Talk: A Framework for Formal Fairness Certificates}

\begin{document}

\twocolumn[
  \icmltitle{Language Models That Walk the Talk: \\ A Framework for Formal Fairness Certificates}



  \icmlsetsymbol{equal}{*}

  \begin{icmlauthorlist}
    \icmlauthor{Danqing Chen}{equal,yyy}
    \icmlauthor{Tobias Ladner}{equal,yyy}
    \icmlauthor{Ahmed Rayen Mhadhbi}{yyy} 
    \icmlauthor{Matthias Althoff}{yyy}
  \end{icmlauthorlist}

  \icmlaffiliation{yyy}{Technical University of Munich, Germany}

  \icmlcorrespondingauthor{Danqing Chen}{chen.danqing@tum.de}
  \icmlcorrespondingauthor{Tobias Ladner}{tobias.ladner@tum.de}

  \icmlkeywords{Machine Learning, ICML}

  \vskip 0.3in
]



\printAffiliationsAndNotice{}  

\begin{abstract}
As large language models become integral to high-stakes applications, ensuring their robustness and fairness is critical. Despite their success, large language models remain vulnerable to adversarial attacks, where small perturbations, such as synonym substitutions, can alter model predictions, posing risks in fairness-critical areas, such as gender bias mitigation, and safety-critical areas, such as toxicity detection. While formal verification has been explored for neural networks, its application to large language models remains limited. This work presents a holistic verification framework to certify the robustness of transformer-based language models, with a focus on ensuring gender fairness and consistent outputs across different gender-related terms. Furthermore, we extend this methodology to toxicity detection, offering formal guarantees that adversarially manipulated toxic inputs are consistently detected and appropriately censored, thereby ensuring the reliability of moderation systems. By formalizing robustness within the embedding space, this work strengthens the reliability of language models in ethical AI deployment and content moderation.
\end{abstract}

\section{Introduction}

As large language models become integral to academia and industry, particularly in high-stakes decision-making and safety-critical domains, ensuring their robustness, and trustworthiness is crucial~\cite{garcia2024trustworthy, huang2024survey}. While their transformer-based architecture with attention mechanisms has driven widespread adoption~\cite{vaswani2017attention}, concerns about robustness and safety persist, especially in tasks like fairness enforcement and content moderation.

A key challenge is their susceptibility to adversarial perturbations~\cite{goodfellow2015explaining, szegedy2013intriguing}, where small perturbations, such as synonym substitutions or paraphrasing, can manipulate model behavior~\cite{alzantot2018generating, behjati2019universal, li2019textbugger, liang2017deep, jin2020bert, hsieh2019robustness}. While such perturbations may be adversarially crafted in settings like toxicity detection~\cite{bespalov2023towards, schmidhuber2024llm}, they can also arise naturally in fairness-critical tasks such as salary prediction, where gender bias may occur, highlighting the need for robust gender bias mitigation~\cite{tumlin2024fairnnv}. Many natural language processing tasks operate in discrete input spaces (language), where verifying robustness by enumerating all synonym substitutions becomes computationally infeasible due to combinatorial explosions. This holds especially for longer text sequences and words with many synonyms, which makes adversarial robustness verification particularly challenging~\cite{bonaert2021fast, huang2019achieving}. Specifically, the number of possible 20-token-long adversarial sequences drawn from a 32k-token (BERT-size) vocabulary is equal to $32\text{k}^{20} \approx 10^{90}$, a figure that exceeds the estimated number of atoms in the observable universe \cite{kumar2024certifying}.

\section{Background}
\label{gen_inst}

\begin{figure*}
    \centering
\includegraphics[width=0.95\linewidth]{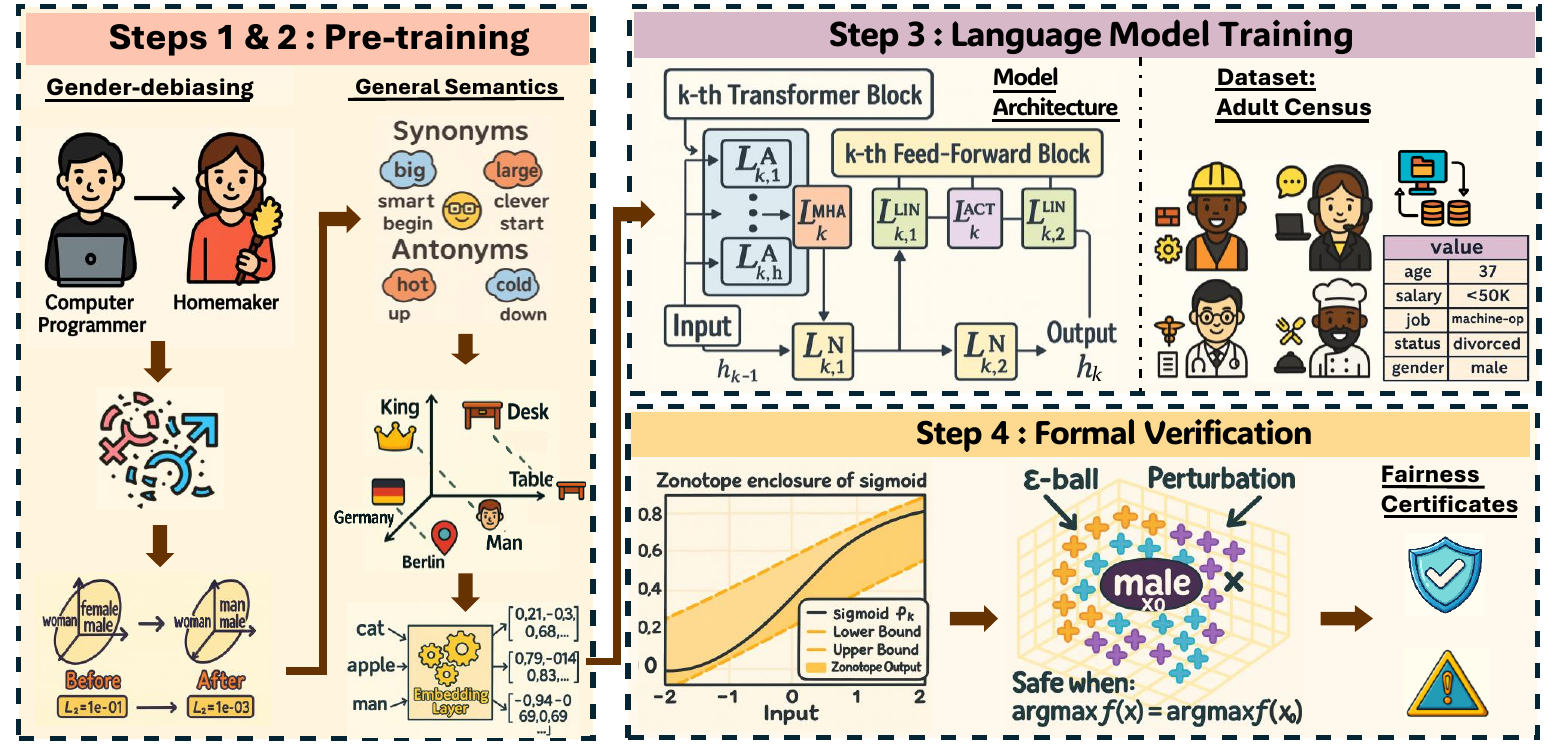}
    \caption{Overview of the framework explained in \Cref{sec:exp_setup} involving pre-training (step~1 and~2), model training (step~3), and formal verification (step~4). The transformer architecture in step 3 is from~\cite{vaswani2017attention}.}
    \label{fig:pipeline}
\end{figure*} 

We consider a general neural network classifier architecture that supports any element-wise nonlinear activation function. Given an input $\nnInput \in \R^n$, the classifier is represented by $\NN : \R^n \rightarrow \R^c$. A variety of tools have been developed to formally verify neural network properties, with adversarial robustness being the most extensively studied aspect~\cite{brix2023yearsinternationalverificationneural}. The verification problem for neural networks can be formally stated as follows:

\paragraph{Neural network verification.}
Let $\NN: \R^n \rightarrow \R^m$ be a neural network with input $\nnInput\in\R^n$ with $\nnOutput=\NN(\nnInput)$. 
Let $\psiin(\nnInput)$ be an input specification and $\psiout(\nnOutput)$ an unsafe output specification.

The verification problem determines if
\begin{equation}
\forall \nnInput \in \R^n\colon \psiin(\nnInput) \implies \neg\psiout(\NN(\nnInput)).
\end{equation}
If there exists a point for which this statement cannot be proven, the network is considered \emph{unsafe}; otherwise, it is considered \emph{safe}.

Exact verification of robustness properties in neural networks is often computationally infeasible~\cite{katz2017reluplex}. 
Therefore, sound but incomplete methods are usually used for larger networks~\cite{katz2019marabou, zhang2018efficient, bak2021nnenum, lopez2023nnv, singh2018fast,duong2024harnessing}. 
In particular, we use zonotopes~\cite{girard2005reachability,bonaert2021fast} to enclose the output of a neural network in this work, which has been shown to be a suitable foundation for analyzing larger neural networks~\cite{singh2018fast, 8418593, mirman2018differentiable}. 
As zonotopes offer an efficient and scalable way to represent perturbed input (or embedding) spaces, they have also been used to verify transformer-based models~\cite{bonaert2021fast}.
A detailed explanation of zonotope-based neural network verification can be found in \Cref{sec:zonotope}.
However, our work is agnostic to the underlying verification engine and others can be used as well~\cite{brix2023fourth,brix2024fifth}.

\section{Methodology}

\label{sec:exp_setup}

This section presents the fairness verification framework in this work. 
As a running example, we illustrate the process using gender fairness verification in a salary prediction task. 
An overview of the framework
is provided in \Cref{fig:pipeline}.

\paragraph{Step 1: Gender-debias pre-training (\Cref{sec:pretraining_gender_debias}).}
We begin by pre-training an embedding layer on curated gendered word pairs, encouraging proximity between gendered pairs while maximizing distance from gender-neutral words. This enforces semantic neutrality and fairness at the embedding level through $\ell_2$ distance. 

\paragraph{Step 2: General semantic pre-training (\Cref{sec:pretraining_gender_debias}).}
To enhance general semantic understanding, we further pre-train the model using synonym/antonym pairs from WordNet, enabling it to capture deeper semantic and syntactic relationships.

\paragraph{Step 3: Language model training (\Cref{sec:lm_training}).}
The pre-trained embedding layer is integrated into a transformer-based language model and kept frozen during downstream training of the model, and then the language model is trained on a fairness-critical task involving gender bias mitigation.

\paragraph{Step 4: Formal verification (\Cref{sec:zonotope_formal}).}

We evaluate fairness and robustness regarding synonym substitutions by perturbing the input embedding space and analyzing the resulting changes in model outputs using formal verification.


\subsection{Pre-training the embedding model}
\label{sec:pretraining_gender_debias}

Since many AI-integrated automated decisionsication) can significantly impact people’s lives, it is paramou (including which individuals will receive jobs, loans, and mednt to assess and improve the ethics of the decisions made by these automated systems~\cite{10.1145/3494672, 10.1145/3457607}. A straightforward approach to promoting fairness in machine learning, defined as the absence of bias and discrimination, is to preclude the use of sensitive attributes during model training and inference~\cite{lepri2018fair,calders2010three,kamiran2010discrimination, barocas2016big}. For instance, in designing a race-blind or gender-blind decision-making system, one might either exclude sensitive attributes, such as gender, from the feature set entirely~\cite{lepri2018fair}, or alternatively, represent them with similar embeddings that capture shared semantic meaning, placing them in the same semantic class to ensure equivalent treatment by the model~\cite{bolukbasi2016man}. The corresponding experiments for both these scenarios are presented in \Cref{sec:result_main}.


Word embedding models map words to vectors based on the assumption that words appearing in similar contexts share similar meanings~\cite{10.1145/3494672, mikolov2013efficient, mikolov2013distributed}. They capture semantic relationships and are widely used in natural language processing tasks~\cite{10.1145/3494672}. However, studies have shown that embeddings can encode biases~\cite{bolukbasi2016man, brunet2019understanding, Caliskan_2017, zhao2018learning}; for instance, \token{computer programmer} may appear closer to \token{male} than \token{female}, raising concerns about potential discrimination in AI-based decision-making systems.

This work employs pre-training-based transfer learning to train an embedding layer from scratch and cluster gender-related words while distancing unrelated ones. This offers greater transparency, controllability, and verifiability, enabling direct mitigation of gender bias during embedding layer training by mapping all gender-related words to a cohesive region in the embedding space. This semantic alignment enables formal verification. Similarly, this framework can also be applied to adversarial substitutions in the context of toxicity detection, ensuring robust censorship of inappropriate content such as profanity, hate speech, and abusive or otherwise harmful language. The approach is motivated by the need to better capture semantic relationships—especially involving gender, synonyms, or toxic terms—that are often missed by backpropagation alone\footnote{Transfer learning is necessary because some embedding layers (e.g., Keras) are randomly initialized~\cite{keras2023}, and training alone often fails to capture complex relationships like gender or toxicity.}. Prior work supports this, showing embeddings trained on task-specific data can outperform static embeddings like Word2Vec~\cite{10.1145/3440840.3440847}.
Similar practices are seen in domains such as fake news detection~\cite{liu2018early}, highlighting the value of transfer learning to improve task performance and propagate semantic knowledge~\cite{9134370}.

\paragraph{Contrastive loss function.}
The effectiveness of machine learning models relies heavily on data representation, with certain feature sets presumed tocapture core characteristics across tasks~\cite{9226466}. A good representation promotes intra-class compactness and inter-class separability~\cite{khosla2020supervised, jaiswal2020survey, haochen2021provable, wang2021understanding}. Contrastive learning has proven highly effective in enhancing representations across NLP, visual computing, and beyond~\cite{jaiswal2020survey, caron2020unsupervised, he2020momentum, chen2020simple, haochen2021provable, saunshi2019theoretical, chen2020big, chen2020improved, he2020momentum, 1467314}. To learn semantically meaningful embeddings, we employ a contrastive loss that separates similar and dissimilar word pairs in the embedding space.

Let $h_1, h_2 \in \R^d$ denote the embeddings of a pair of input words, and let $d = \lVert h_1 - h_2 \rVert_2$ be the predicted Euclidean distance between them. Given a binary label $y \in \{0, 1\}$ indicating whether the pair is dissimilar ($y = 1$) or similar ($y = 0$), the contrastive loss is defined as:
 \begin{equation}
 \mathcal{L}(y, d) = 
 (1 - y) \cdot \left( \alpha  d \right) 
 + y \cdot \left( \max(0, m - d) \right).
 \label{eq:enhanced_contrastive_loss}
 \end{equation}
The loss function used in this work is inspired by the formulation introduced in \cite{1467314} and has been adapted to suit our specific task. Here, the predefined margin $m > 0$ specifies the minimum required separation between dissimilar pairs, while the hyperparameter $\alpha \geq 0$ regulates the influence of an additional linear attraction term applied to similar pairs. The loss encourages the model to minimize distances between similar pairs while enforcing a separation of at least $m$ units between dissimilar pairs in the embedding space. Examples of gender-related terms are listed in \Cref{sec:gender-specific-dataset}, with contrastive loss details in \Cref{sec:contras_loss_de} and word pair construction in \Cref{sec:data_cons_pre}. Additional pre-training with synonym and antonym pairs uses WordNet's Synset structure~\cite{10.1145/219717.219748} to model semantic similarity and contrast.

\subsection{Training the language model}
\label{sec:lm_training}

The pre-trained embedding layer is subsequently integrated into a transformer-based language model, with the architecture proposed in~\cite{vaswani2017attention} and kept frozen and not updated via backpropagation during training to preserve linguistic knowledge, retaining general linguistic knowledge captured during pre-training, following common practice in models such as ELMo and ULMFiT~\cite{peters2018deepcontextualizedwordrepresentations,howard2018universallanguagemodelfinetuning}.

\subsection{Verifying the fairness of language models}
\label{sec:zonotope_formal}

To evaluate the robustness and fairness of models, we use formal verification based on zonotopes (\Cref{sec:zonotope}). 
Given a sentence, we want to verify that the prediction of the model stays the same even if the words are replaced by synonyms or gender-related terms.
Please note that a brute-force technique quickly becomes computationally infeasible due to combinatorial explosions~\cite{kumar2024certifying}, particularly for longer sentences and words with many synonyms.

\paragraph{Capturing synonyms with perturbation balls.}
Thus, we capture synonyms (or gender-related or equivalent toxic terms) by modeling $\ell_\infty$-balls around each word representation in the embedding space as zonotopes. 
If these perturbation balls are large enough, they capture the synonyms as we pre-trained the embedding space such that these are in close proximity (\Cref{sec:pretraining_gender_debias}).
A fair model should maintain consistent predictions under sufficiently large perturbations.

We perturb the embedding of each word in a sentence by some perturbation radius $\nnPertRadius>0$.
We then employ binary search to find the largest $\nnPertRadius_\text{max}$ for which we can still verify that the prediction remains unchanged.
Once $\nnPertRadius_\text{max}$ is obtained, we assess if all synonyms of the words of interest lie within the computed perturbation ball.
This can be easily checked by determining a threshold $\maxDistance$ given by the largest $\ell_\infty$ distance between each word $i$ and its respective synonyms in the embedding space.
Let $\nnEmbedding_i$ be the embedding of word $i$, then
\begin{equation}
    \maxDistance = \max_i \max_{j\in\mathcal{N}(i)} \norm{\nnEmbedding_i - \nnEmbedding_j}_\infty,
\end{equation}
where $\mathcal{N}(i)$ contains all synonyms and related words of word $i$.
If $\maxDistance \leq \nnPertRadius_\text{max}$ holds for all words in the sentence, 
we know by the verification result that any possible combination of synonyms does not change the prediction. 
\begin{algorithm}[t]
\caption{Formal Fairness Verification}
\label{alg:synonym_verification}
\begin{algorithmic}[1]
\REQUIRE Dataset $\{S_1, \dots, S_N\}$, model $\NN$, embedding $\nnEmbedding$, synonym sets $\mathcal{N}(i)$ for each word $i$
\ENSURE Fairness score $\fairnessScore \in [0,1]$

\STATE $c \gets 0$ \hfill // count of certified sentences

\FOR{$l = 1$ to $N$}
    \STATE \text{$\nnPertRadius_{\text{max},l} \gets$ find max. verifiable perturbation}
    \STATE $\maxDistance_l \gets \max_{i \in S_l} \max_{j \in \mathcal{N}(i)} \norm{\nnEmbedding(i) - \nnEmbedding(w_j)}_\infty$
    \IF{$\maxDistance_l \leq \nnPertRadius_{\text{max},l}$}
        \STATE $c \gets c + 1$
    \ENDIF
\ENDFOR

\STATE $\fairnessScore \gets \frac{c}{N}$
\STATE \textbf{return} $\fairnessScore$
\end{algorithmic}
\end{algorithm}

This verification setup returns a formal fairness certificate:
Specifically, we construct input perturbation sets that include all fairness-sensitive terms, including different gender terms or different paraphrasing of toxic language, and verify that no variation (of gender attributes) or substitution (of semantically equivalent toxic terms) within the entire perturbation set can alter the prediction of the model. Fairness is then certified if no such perturbation exists.

\paragraph{Fairness score.}
We can assess the overall fairness of a model by checking the percentage of all of the sentences in our test dataset, such a certificate can be obtained.
In particular, let there be $N$ sentences in a held-out test dataset (not seen during model training), then the fairness score $\fairnessScore$ can be computed by
\begin{equation}
    \label{eq:fairness-score}
    \fairnessScore = \frac{1}{N}\sum_{l=1}^N \mathbf{1}_{(\maxDistance_l \leq \nnPertRadius_\text{max,l})},
\end{equation}
where $\mathbf{1}_{(\cdot)}$ is an indicator if the certificate is obtained for sentence $l$.
Trivially, A higher $\Psi$ (closer to 1) implies a fairer model. The whole verification procedure is shown in~Alg. \ref{alg:synonym_verification}.

\section{Experimental Results}
\raggedbottom

\subsection{Datasets}

\paragraph{Gender bias mitigation.}


To mitigate gender bias in the learned representations, we employ transfer learning to pre-train a Keras embedding layer using a curated set of gender-specific words curated in prior work~\cite{bolukbasi2016man}. The model is trained to cluster gendered word pairs (e.g., \token{man}–\token{woman}, \token{male}–\token{female}, \token{king}–\token{queen}) into a shared semantic class, minimizing the distance between them in the embedding space, while ensuring that unrelated words are separated. This strategy helps the model internalize semantic neutrality with respect to gender, which is critical for downstream fairness-sensitive tasks. Detailed preprocessing steps on the pre-training dataset can be found in \Cref{sec:adult_preprocessing}.

The dataset used in the model training for the task of gender bias mitigation in this study is the Adult Census dataset~\cite{adult_2}, including demographic information from the 1994 U.S. Census. The task is to predict whether an individual’s income exceeds \$50,000 per year, revealing potential biases in predictions based on race, gender, and other sensitive features. This dataset has been widely used in the context of fairness in machine learning~\cite{lahoti2020fairness, friedler2019comparative, patel2024design, le2022survey}. Detailed preprocessing steps, including the serialization of tabular data by  converting to a sentence dataset so that it is consumable by transformers \cite{badaro2023transformers}, with prior literature supporting this method \cite{chen2019tabfact,suadaa2021towards,chen2020logical}, can be found in \Cref{sec:data_cons_pre_tab}.

\paragraph{Toxicity detection.}
The prevalence of toxic comments on social networking sites poses a significant threat to the psychological well-being of online users, which is why moderation is crucial~\cite{pavlopoulos2020toxicity,martens2015toxicity}. To address this challenge, researchers have turned to machine learning algorithms as a means of categorizing and identifying toxic contents~\cite{patel2025detecting, 10.1145/3038912.3052591}.

The pre-training of embedding layers follows the same framework as that of the gender bias mitigation experiment, using a bad words collection from Google’s publicly available bad words list\footnote{\url{https://code.google.com/archive/p/badwordslist/downloads}}. For the training and evaluation of the language model, the Jigsaw Toxic Comment Classification dataset~\cite{jigsaw2017toxic} was utilized. This dataset has been widely adopted in various studies~\cite{patel2025detecting,chakrabarty2020machine,revelo2023classification,marchiori2022bias,gupta2022machine}. In this study, we use a curated version of the Jigsaw toxic comments dataset~\cite{hateoffensive} 
designed to evaluate the severity of toxic language by incorporating annotator agreement probabilities across three categories: neutral, offensive, and hate speech. We formulated a binary classification task, where neutral comments are labeled as non-toxic (label = 0), and both offensive and hate speech comments are labeled as toxic (label = 1).
\raggedbottom

\subsection{Main results and discussions}

\label{sec:result_main}
In this section, we present the results of our experiments. We evaluated our approach on text classification language models of up to 10 blocks of transformers. We provide details on the pre-training hyperparameters in \Cref{sec:hyperparaneter_tuning} and ablation studies, including more models in \Cref{sec:ablation-study}.

\begin{figure}[ht]
\centering
\begin{minipage}{0.45\linewidth}
    \centering
    \includegraphics[width=\linewidth]{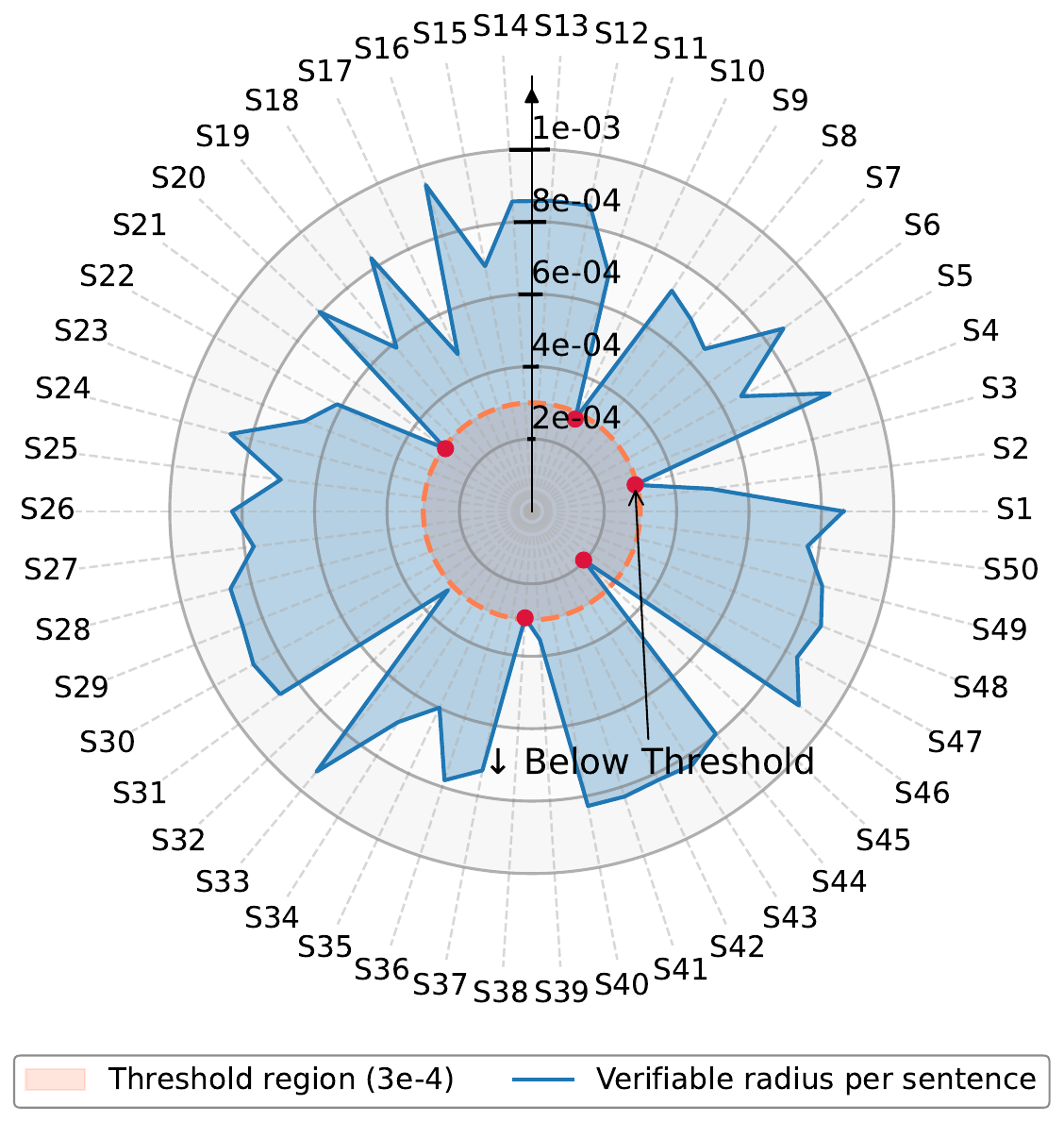}
    \caption*{(a) 8-block model.}
\end{minipage}
\hfill
\begin{minipage}{0.45\linewidth}
    \centering
    \includegraphics[width=\linewidth]{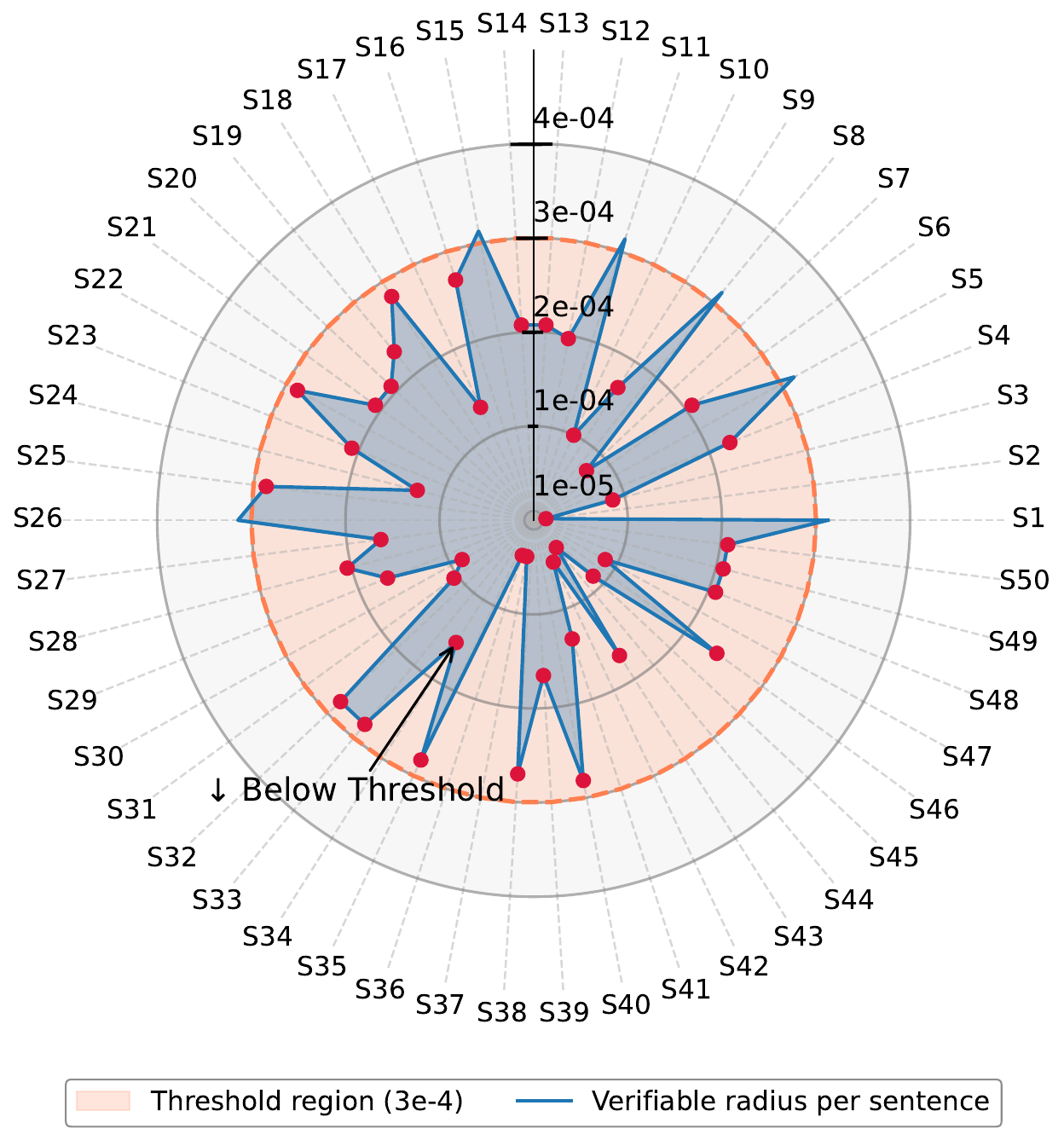}
    \caption*{(b) 10-block model.}
\end{minipage}
\caption{Verifiability radar plot with gender-distance reference circle (coral dashed circle) in gender bias mitigation with 8 and 10 blocks of transformers. The axes S1 to S50 stand for the 50 sentences from the held-out test dataset.}
    \label{fig:verifiability_radar_8_10}
\end{figure}

\begin{table}[ht]
\centering
\small  
\setlength{\tabcolsep}{4pt}  
\renewcommand{\arraystretch}{1.1}  
\caption{Model performance (test accuracy) and verification metrics. 'NA' (Not Applicable) indicates that gender terms were omitted; hence, fairness could not be computed.}
\begin{tabular}{lcc}
\toprule
 & \multicolumn{2}{c}{\cellcolor{gray!10}\textbf{Salary Prediction \& Gender Bias Mitigation}} \\
\cmidrule(lr){2-3}
\textbf{Model} & \textbf{Test Acc. ($\uparrow$)} & \textbf{Fairness Score $\fairnessScore$ ($\uparrow$)} \\
\midrule
\multicolumn{3}{l}{\textit{Gender attributes excluded}} \\
8 blocks & 82\% & NA \\
10 blocks & 83\% & NA \\
\midrule
\multicolumn{3}{l}{\textit{Gender attributes included}} \\
8 blocks & 88\% & 90\% \\
10 blocks & 88\% & 12\% \\
\midrule
 & \multicolumn{2}{c}{\cellcolor{gray!10}\textbf{Toxicity Detection \& Content Moderation}} \\
\cmidrule(lr){2-3}
\textbf{Model} & \textbf{Test Acc. ($\uparrow$)} & \textbf{Fairness Score $\fairnessScore$ ($\uparrow$)} \\
\midrule
3 blocks & 95\% & 50\%\\
\bottomrule
\end{tabular}

\label{tab:gender_toxicity_result}
\end{table}

\paragraph{Gender fairness.}
We evaluate the fairness of our models using the fairness score $\fairnessScore$ from eq~\eqref{eq:fairness-score}. To compute the fairness score, we constructed the test dataset by randomly selecting 50 sentences—25 from each class (label = 0 or 1 for whether the salary exceeds \$50,000)—from the training set. This represents the model's overall fairness. Gender-related tokens are identified as the 50 closest neighbors to the word \token{male} and \token{female} (the primary gender indicators in the Adult Census dataset) in the embedding space.
The largest $\ell_\infty$ distance between any of their nearest 50 neighbors is $3 \times 10^{-4}$ (for both the word \token{male} and \token{female}. Notably, \token{male} and \token{female} appear within each other's 50 nearest neighbors, demonstrating the effectiveness of pre-training). We use this value as a reference threshold (illustrated by the coral circle in \Cref{fig:verifiability_radar_8_10}). Tokens within this threshold are treated as semantically equivalent (see \Cref{tab:embedding_distances_gender}).
A full list of gender-related neighbors is provided in \Cref{sec:distance_complete}. \Cref{tab:gender_toxicity_result} presents the model’s test accuracy and fairness scores after training. Test accuracy is a crucial component of our evaluation, as fairness assessments are only meaningful when applied to models that demonstrate reliable predictive performance. \Cref{fig:verifiability_radar_8_10} shows radar plots of per-sentence verifiability, where each axis represents a sentence and the blue region’s tip indicates the maximum verifiable perturbation radius $\nnPertRadius_\text{max}$. Sentences with tips outside the coral dashed circle (radius $\maxDistance$) are deemed verifiable (fair), while red dots mark unverifiable cases.

 The 8-block language model achieved a fairness score of 90\%, indicating that its predictions remain robust and unbiased with respect to gender across all tested scenarios in the salary prediction task involving demographic attributes. We are noting that the fairness scores of the model with a deeper architecture decreased compared with smaller models (10 vs. 8 transformer blocks), despite exhibiting similar test accuracy. This decline might be due to larger models being harder to verify~\cite{katz2017reluplex} or the model indeed being unfair. 
 To this date, we treat both cases the same way and only count models as fair if they are verifiably so.
 However, it is important to emphasize that the primary objective of this work is not solely to ensure that every evaluated model is verified as fair or safe, 
but rather to establish a reliable framework for assessing these properties. 
Ideally, our framework exposes adversarial examples and biases and can guide the development of fairer systems, such as through pre-training strategies that cluster gendered tokens into the same semantic class on the embedding layers utilized in the model. Thus, when the verification results reveal unfairness or potential unfairness, this indicates that the model might need further training towards fairer predictions.

\begin{table}[ht]
\centering
\small  
\setlength{\tabcolsep}{4pt}  
\renewcommand{\arraystretch}{1.1} 
\caption{Example $\ell_\infty$ distances between some common gender-related word embeddings.}
\begin{tabular}{r c l c}
\toprule
\multicolumn{3}{c}{\textbf{Word pair}} & $\ell_\infty$ \textbf{Distance} \\
\midrule
\token{female} & $\leftrightarrow$ & \token{male}   & $3\text{e}{-04}$ \\
\token{female} & $\leftrightarrow$ & \token{man}    & $3\text{e}{-04}$ \\
\token{male}   & $\leftrightarrow$ & \token{woman}  & $2\text{e}{-04}$ \\
\token{man}    & $\leftrightarrow$ & \token{woman}  & $3\text{e}{-04}$ \\
\token{he}     & $\leftrightarrow$ & \token{she}    & $2\text{e}{-04}$ \\
\bottomrule
\end{tabular}

\label{tab:embedding_distances_gender}
\end{table}

 While removing gender attributes from the input may seem like a straightforward way to mitigate bias, this approach is often insufficient in practice. A prior study \cite{8987497} shows that removing sensitive attributes like gender does not always lead to a fairer model, as models can still learn bias from correlated features—and may even become more biased by relying on proxy attributes. Similar studies support this, showing that simply removing sensitive attributes is insufficient, as protected features may be encoded or correlated with other variables \cite{loi2021transparency, gonzalez2024review, goethals2024precof}. Our experimental result (\Cref{tab:gender_toxicity_result}) further shows that excluding gender-related information in both the pre-training and language model training phase leads to a drop in model accuracy, indicating that such features can play a legitimate role in prediction, highlighting the necessity of our proposed fairness score to assess and manage potential bias rather than outright exclusion.

\begin{figure}[ht]
\centering
\includegraphics[width=0.7\linewidth]{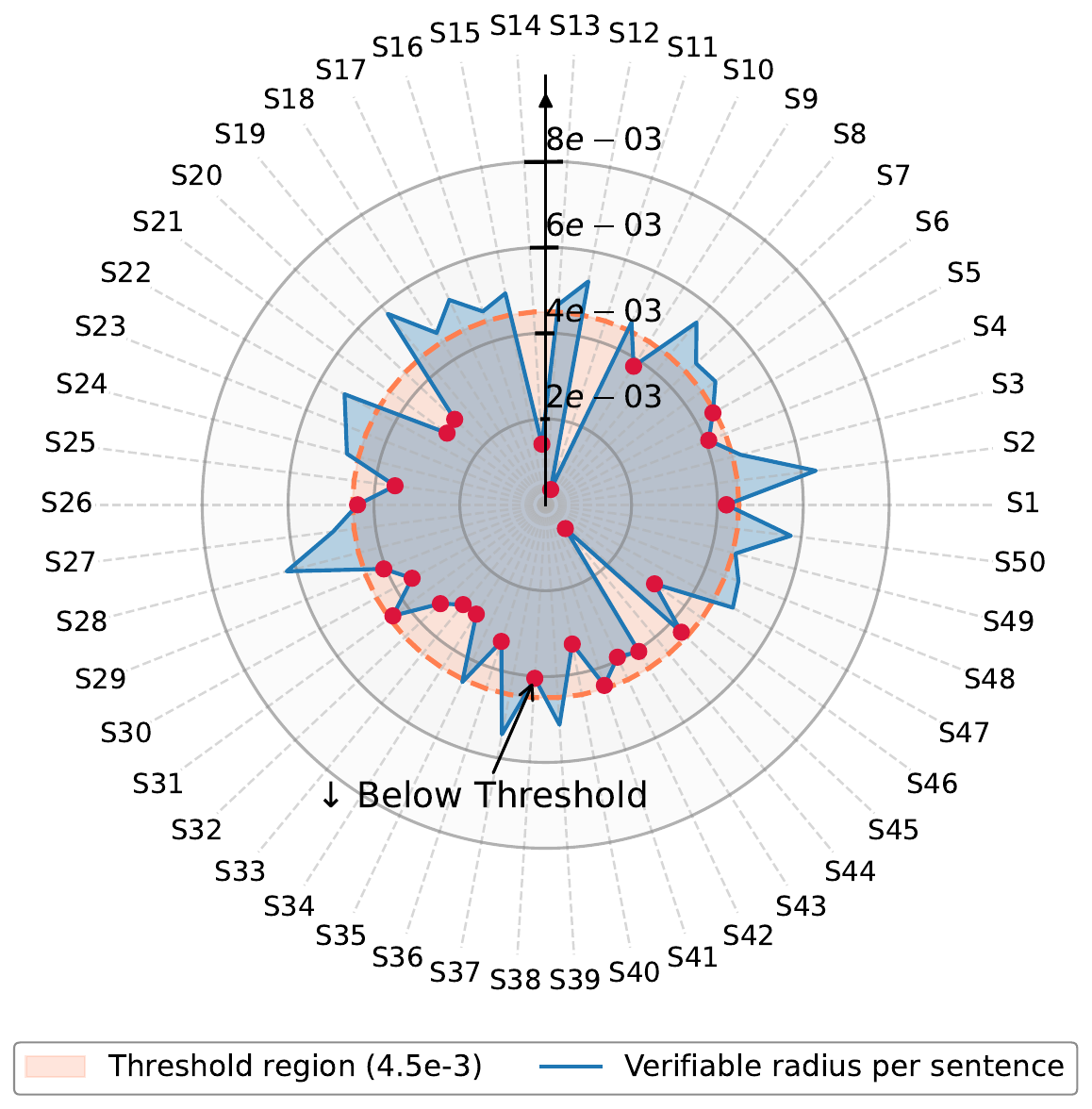}
\caption{Verifiability radar plot with toxicity-distance reference circle (coral dashed circle) for the 3-block toxicity detection model in content moderation. The axes S1 to S50 stand for the 50 sentences from the held-out test dataset.}
\label{fig:toxity_3b}
\end{figure}

\paragraph{Toxicity detection.}

For the toxicity detection experiment, we begin by identifying a set of common toxicity-related terms from the dataset. Using a pre-trained embedding space, we then retrieve the 50 nearest neighbors for each term, following the same approach used for gendered terms. For each toxic word, we compute the maximum $\ell_\infty$ distance among its 50 nearest neighbors, representing the semantic boundary for meaning-preserving substitutions. The largest of these distances defines the reference circle (coral, in \Cref{fig:toxity_3b}), used to assess whether the model's predictions remain stable under semantically equivalent toxic term replacements. An example of the full neighbor set is provided in \Cref{sec:distance_complete}.


\begin{table}[ht]
\centering
\small 
\setlength{\tabcolsep}{4pt}
\renewcommand{\arraystretch}{1.1}
\caption{Example $\ell_\infty$ distances between selected toxic terms and their nearest neighbors in embedding space. \textbf{The vocabulary is publicly available via a Google-hosted archive dataset and does not reflect the opinions of the authors}.}
\begin{tabular}{r c l c}

\toprule
\multicolumn{3}{c}{\textbf{Word pair}} & $\ell_\infty$ \textbf{Distance} \\
\midrule
\token{mother-fucker} & $\leftrightarrow$ & \token{\#\#ore}   & $3.4\text{e}{-03}$ \\
\token{mother-fucker} & $\leftrightarrow$ & \token{tits}      & $4.0\text{e}{-03}$ \\
\token{mother-fucker} & $\leftrightarrow$ & \token{fuck}      & $4.1\text{e}{-03}$ \\
\token{bitch}         & $\leftrightarrow$ & \token{cock}      & $2.6\text{e}{-03}$ \\
\token{bitch}         & $\leftrightarrow$ & \token{ho}        & $4.0\text{e}{-03}$ \\
\token{gay}           & $\leftrightarrow$ & \token{ass}       & $3.3\text{e}{-03}$ \\
\token{gay}           & $\leftrightarrow$ & \token{\#\#ker}    & $4.0\text{e}{-03}$ \\
\token{gay}           & $\leftrightarrow$ & \token{bi}        & $2.4\text{e}{-03}$ \\
\token{nigger}        & $\leftrightarrow$ & \token{monkey}    & $3.6\text{e}{-03}$ \\
\token{nigger}        & $\leftrightarrow$ & \token{\#\#er}     & $3.4\text{e}{-03}$ \\
\token{nigger}        & $\leftrightarrow$ & \token{cock}      & $2.6\text{e}{-03}$ \\
\bottomrule
\end{tabular}

\label{tab:toxicity_neighbors_clean}
\end{table}

The per-sentence verifiability results for the 3-block transformer model used in toxicity detection are shown in \Cref{fig:toxity_3b}. The model achieved a test accuracy of 95\% and a fairness score of 50\% after training and formal verification. This indicates that the model can reliably censor toxic content—regardless of any adversarial paraphrasing using toxic equivalents—in 50\% of the tested cases. In addition, \Cref{tab:toxicity_neighbors_clean} presents the pairwise 
$\ell_\infty$  distances between frequently occurring toxicity-related word pairs in the Jigsaw dataset and throughout verification. The majority of them fall within the maximum verifiable noise radius for every sentence. This is particularly important in the context of content moderation. For example, an Internet user might attempt to evade detection by slightly modifying toxic terms (e.g., changing \token{arse} to \token{azzhole}). However, a model that has obtained a fairness certificate would guarantee to be able to verify such edits and maintain detection capability, demonstrating the effectiveness of our verification approach in real-world scenarios such as toxicity detection and content moderation\footnote{The authors used a smaller transformer for toxicity detection, as explicit cues make the task simpler and larger models led to overfitting, as observed empirically.}.

To further substantiate the effectiveness of our framework, we include an ablation study in \Cref{sec:ablation-study} that compares the fairness scores of normally trained and adversarially trained models. The results of this ablation study provide additional evidence of the framework’s ability to distinguish and reward models that incorporate adversarial training.

\section{Related work}\label{sec:related-work} 
This work continues the progressive effort to formally verify neural networks in recent years~\citep{brix2023fourth}: As neural networks are vulnerable to adversarial attacks~\citep{goodfellow2015explaining}, formally verifying them against input perturbations is of immense importance in safety-critical scenarios. Neural network verifiers can generally be categorized into optimization-based verifiers~\cite{zhang2018efficient, katz2019marabou, henriksen2020efficient, singh2019abstract} and approaches using set-based computing~\cite{gehr2018ai2, singh2018fast,lopez2023nnv,kochdumper2022open}. Early approaches focused on computing the exact output set of neural networks with ReLU activations, which has been shown to be NP-hard~\cite{katz2017reluplex}. Convex relaxations of the problem~\cite{singh2018fast,xu2020fast} and efficient branch-and-bound techniques~\cite{wang2021beta}
have shown to achieve impressive verification results~\cite{brix2023fourth}.
Recently, non-convex relaxations have also been shown to be effective in this field of research~\cite{kochdumper2022open,wang2023polar,ladner2023automatic,fatnassi2023bern,ortiz2023hybrid}.

Many works have focused on verifying the safety and fairness of neural networks. For example, FairNNV~\cite{tumlin2024fairnnv} verifies individual and counterfactual fairness via reachability analysis, while safety in robotic systems has been studied using formal methods~\cite{sun2019formal}. Probabilistic approaches, such as those using Markov Chains~\cite{sun2021probabilistic}, FairSquare~\cite{10.1145/3133904}, and VeriFair~\cite{10.1145/3360544}, enable rigorous group fairness verification. The OVERT algorithm~\cite{sidrane2022overt} enables sound safety verification of nonlinear, discrete-time, closed-loop dynamical systems with neural network control policies. Discriminatory input detection and fairness analysis are addressed by DeepGemini~\cite{xie2023deepgemini}, complemented by Fairify’s SMT-based method for verifying individual fairness~\cite{biswas2023fairify}. An automated verification framework for feed-forward multi-layer neural networks based on Satisfiability Modulo Theories (SMT) is proposed in~\cite{huang2017safety}, while NNV~\cite{tran2020nnv} offers a set-based verification approach tailored for deep neural networks. Other methods include training-based fairness improvement~\cite{borcatasciuc2022provablefairnessneuralnetwork}, certified representation learning~\cite{ruoss2020learning}, structured data verification~\cite{john2020verifying}, global fairness verification with CertiFair~\cite{khedr2023certifair}, parallel static analysis for causal fairness~\cite{urban2020perfectly}, and Bayesian and satisfiability-based tools like FVGM and Justicia~\cite{ghosh2021justicia, ghosh2022algorithmic}. A new approach to measuring and addressing discrimination in supervised learning, where a model is trained to predict an outcome based on input features~\cite{hardt2016equality}. 
Prior work has predominantly focused on fairness verification of standard feed-forward neural networks, whereas our approach focuses on verifying both the fairness and safety properties of transformer-based language models.

Bias evaluation metrics for large language models can be categorized by what they use from the model \cite{gallegos2024bias}, such as the embeddings \cite{caliskan2017semantics,dev2020measuring,guo2021detecting}, model-assigned probabilities \cite{webster2020measuring,kurita-etal-2019-measuring,salazar-etal-2020-masked,wang-cho-2019-bert,nadeem-etal-2021-stereoset,kaneko2022unmasking} or generated text \cite{rajpurkar-etal-2016-squad,bordia-bowman-2019-identifying,liang2023holistic,gehman-etal-2020-realtoxicityprompts,sicilia-alikhani-2023-learning,nozza-etal-2021-honest,10.1145/3442188.3445924}. Here, we focus on the generated text, as it is the primary focus of our study. Language models’ responses to prompts can reveal social biases \cite{cheng-etal-2023-marked}, using tasks like prompt completion, conversation, and analogical reasoning~\cite{li2024surveyfairnesslargelanguage}. The standard method involves prompting the model and evaluating its continuation for bias \cite{gallegos2024bias}. 
GPT-4 has been shown to generate more racially stereotyped personas than human writers when prompted to describe demographic groups \cite{cheng-etal-2023-marked}. 
Similarly, earlier work validated gender bias in GPT-3's output using 800 prompts (e.g., ``[He] was very'' vs.\ ``[She] was very''), 
revealing stereotypes in the model output and highlighting the need for debiasing  \cite{brown2020language,10.1145/3461702.3462624}. 
Other analyses show that language models frequently associate certain occupations with specific genders based on co-occurrence patterns \cite{liang2023holistic}. 
The fairness of a large language model is also sometimes evaluated using a second, more powerful, large language model~\cite{eloundou2024first}, which does not provide formal guarantees.
Rather, instead of finding the vulnerability of the first model, it can try to fool the second model to ``pass'' the fairness test~\cite{wang-etal-2024-large-language-models-fair}. 
 
While the vulnerability of large language models to adversarial prompts has been explored~\cite{sun2024trustllm}, research on the non-existence of such prompts remains limited, and the formal verification of transformer-based language models remains largely unexplored. 
As with standard neural networks, the method proposed by~\cite{shi2020robustness} verifies robustness using linear bounds, allowing for at most two perturbed words. This work has been extended to the zonotope domain~\citep{bonaert2021fast}, enabling the verification of longer sentences with all words being perturbed. Other works also investigated finding better bounds of the softmax function within transformers~\citep{wei2023convex,shi2024neural} and investigated the preservation of dependencies between attention heads~\citep{zhang2024galileo}.

Existing efforts—including this work—focus on perturbations applied directly to the embedding space of the input in large language models~\cite{vaswani2017attention}. 
Please note that most existing works apply arbitrary $\epsilon$-perturbations to the input and examine whether such changes can alter the predictions of language models. However, they often overlook a critical assumption: Whether semantically meaningful changes, such as synonym substitution or variation across gender or toxicity-related terms, are actually captured by the $\epsilon$-ball. When synonyms are considered, the focus tends to be limited to traditional definitions, without exploring how the embedding space could be shaped to support other important criteria, such as mitigating gender bias or detecting adversarially crafted yet semantically equivalent toxic terms. In this work, we address this gap by offering a holistic perspective on the formal verification of language models.

\section{Limitation}

Given the scale and complexity of modern large language models~\cite{achiam2024gpt4}, existing verification techniques face a fundamental trade-off between precision and computational cost, rendering them impractical for large-scale models~\cite{shi2020robustness, bonaert2021fast}. Consequently, we exclude larger pre-trained models such as BERT from our evaluation due to their verification intractability. 

This work focuses on gender as a demographic attribute in salary prediction, but the proposed fairness evaluation framework is attribute-agnostic and can be extended to other variables, such as race. While our experiments address binary classification, the framework is, in principle, applicable to multi-class tasks and broader fairness assessments.

Furthermore, our study uses a single state-of-the-art verification method based on zonotopes, applied exclusively to binary text classification models.
This method is sound but incomplete; thus, while an obtained fairness certificate always holds, the method may fail to verify models that are, in fact, fair.
Nonetheless, our verification framework is agnostic to the underlying verification method, and we hope that with the continuous progress in the field of neural network verification~\cite{brix2023yearsinternationalverificationneural},
the fairness of more complex architectures can be verified using our framework.
  
We also want to emphasize that training the embedding layer such that synonyms (or gender-related or toxicity-related terms) are close to each other requires expert knowledge, as such lists must be curated and adapted over time.
Readers should note that this study does not aim to build an exhaustive list of synonyms, bad words, or a collection of all gender-related terms. Rather, it demonstrates that pre-training-based transfer learning enables a simple embedding model to effectively cluster synonyms, gender-related, or toxicity-related words in the embedding space.

\section{Conclusion}

We present a generalized verification framework for formally assessing the fairness and safety of language models and demonstrate it on models with up to 10 transformer blocks.
In particular, the framework consists of carefully designing the embedding space to capture synonyms, gender-related terms, or toxicity-related terms through perturbation using pre-training-based transfer learning.
This embedding layer is then integrated into the custom transformer models for downstream language model training, ensuring the fairness of their predictions.
Lastly, the fairness of the models is formally verified using an approach based on zonotopes.
Importantly, this formal verification technique circumvents the combinatorial complexity of testing individual word substitutions or synonyms by enabling sentence-level verification.

Using this holistic framework, we made sure that gender-related terms are embedded into sufficient proximity through targeted pre-training, allowing different gender terms to be treated as semantically equivalent during inference. This embedding structure is critical in enabling formal verification of gender fairness, as model outputs remained invariant under gender word substitutions. Additionally, we applied the same approach to toxicity-related terms, showing that semantically similar toxic words clustered closely and were consistently detected, even under adversarial perturbations. This highlights the robustness of the system in content moderation tasks. As a result, the framework achieved a fairness score of 90\% for an 8-block transformer language model in the salary prediction task, and 50\% for a 3-block toxicity detection model.
Notably, such results cannot be obtained if the embedding space is not properly designed. These findings underscore the potential utility of our framework in gender bias mitigation for fairness enforcement and its reliability in supporting content moderation systems. 

\section*{Acknowledgements}
This work was supported by the project FAI (No. 286525601) funded by the German Research Foundation (Deutsche Forschungsgemeinschaft, DFG).

\bibliography{custom}
\bibliographystyle{icml2026}

\newpage

\appendix
{\LARGE\bf Appendix}
\setcounter{secnumdepth}{2}
\renewcommand\thesection{\Alph{section}}
\renewcommand\thesubsection{\thesection.\arabic{subsection}}

\section{Zonotope-based verification of neural networks}
\label{sec:zonotope}

\paragraph{Definition.}
Given a center vector $c \in \R^n$ and a generator matrix $G \in \R^{n \times q}$, a zonotope~\cite{girard2005reachability} is defined as
\begin{equation}
    \mathcal{Z} = \shortZ{c}{G} = \left\{ c + \sum_{j=1}^{q} \beta_j G_{(:,j)} \ \middle| \ \beta_j \in [-1, 1] \right\}.
\end{equation}

Zonotopes are convex, centrally symmetric polytopes that are suitable for the formal verification of neural networks due to their compact representation and efficient computation of the required operation.
Generally, the verification process follows these steps:

\begin{enumerate}[label=\arabic*.]
    \item \textbf{Initialization:} Represent the input as a zonotope based on bounds or perturbation limits.
    \item \textbf{Propagation through network:} Propagation of the input set through all layers while ensuring that the output of each layer is properly enclosed.
    \item \textbf{Safety specification checking:} Check if the resulting zonotope satisfies a given safety property.
\end{enumerate}

We explain these steps for a standard feed-forward neural network.
More complex architectures can be verified analogously.

\paragraph{Initialization.}
The input set $\nnInputSet$ should capture all points $\nnInput$ for which $\psiin(\nnInput)$ holds.
Typically, $\nnInputSet$ captures the \( \ell_\infty \)-bounded perturbations around some given input \( \nnInput_0 \in \R^n \).
Given a perturbation radius \( \nnPertRadius \in \R_+ \), the input zonotope is given by
\begin{equation}
\nnInputSet = \shortZ{\nnInput}{\nnPertRadius I_n} \subset \R^n.
\end{equation}

\paragraph{Propagation through network.}
A layer $k$ of a neural network $\NN$ is typically either linear, applying an affine map using a weight matrix $W_k$ and bias $b_k$, or a nonlinear activation $\phi_k$, e.g., ReLU or sigmoid.
Let $\nnHiddenSet_{k-1}=\shortZ{c}{G}$ be the input set to a layer $k$ (with $\nnHiddenSet_0 = \nnInputSet$).
The output of the affine transformation  can be computed exactly using zonotopes and is given by
\begin{equation}
    \nnHiddenSet_{k} = W_k \nnHiddenSet_{k-1} + b_k = \shortZ{W_k c + b_k}{W_k G}.
\end{equation}

\begin{figure*}
   
    \includegraphics[width=1.0\linewidth]{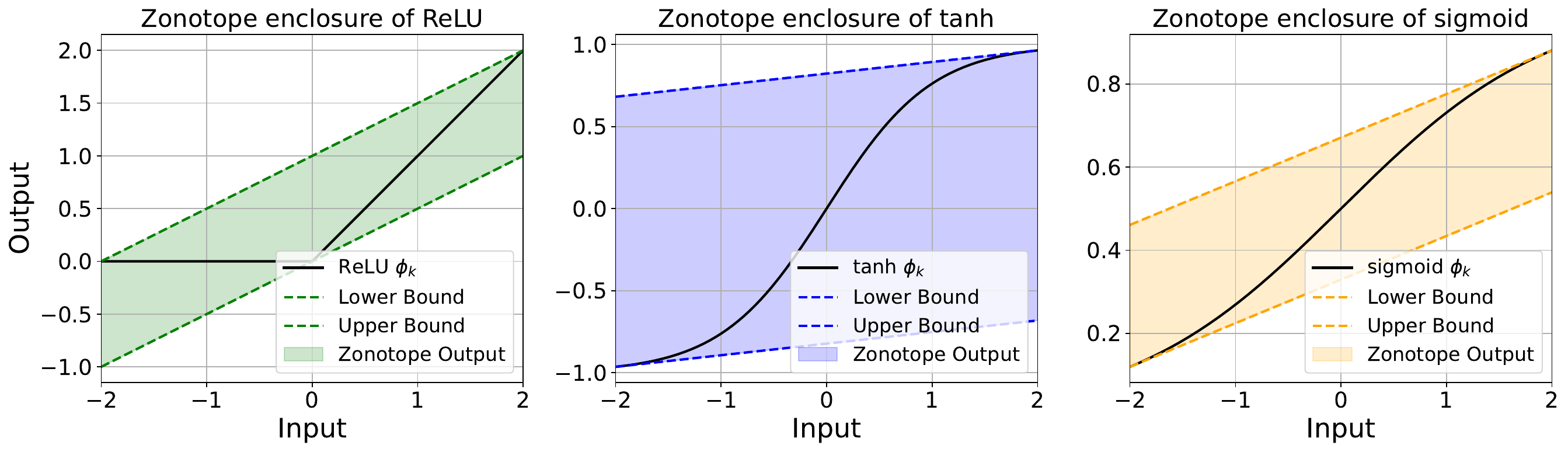}
    \caption{
        Zonotopic enclosure of some of the activation functions $\phi_k$.
        }
    \label{fig:zonotope_example_figure}
\end{figure*}

Nonlinear activations can be computed exactly using zonotopes and are therefore over-approximated to maintain soundness.
Examples of this procedure for different activation functions $\phi_k$ can be found in \Cref{fig:zonotope_example_figure}.

After the output of the last layer $\numLayers$ is enclosed, the output of the neural network is given by $\nnOutputSet=\nnHiddenSet_\numLayers$.

\paragraph{Safety specification checking.}

After the output zonotope $\nnOutputSet$ is obtained, we check whether $\nnOutputSet$ does not intersect with some unsafe set $\unsafeSet$,
where $\unsafeSet$ captures the unsafe specifications $\psiout$.
For classification tasks, it holds that
\begin{equation}
    \nnOutputSet\cap\unsafeSet=\emptyset \implies \forall \nnInput \in \nnInputSet\colon \arg\max \NN(\nnInput) = \arg\max \NN(\nnInput_0).
\end{equation}
Thus, the specifications are formally verified if the emptiness of the intersection $\nnOutputSet\cap\unsafeSet$ is shown.

\section{Dataset construction and preprocessing}
\label{sec:adult_preprocessing}

\subsection{Gender-specific dataset}
\label{sec:gender-specific-dataset}

The gender-specific word list used for pre-training is sourced from~\cite{bolukbasi2016man}, where the authors derived gendered terms from WordNet based on definitional criteria. We adopt this word set and apply the Word Embedding Association Test (WEAT) to categorize terms into \token{male}- and \token{female}-associated groups used predefined attribute set (\Cref{tab:attribute_set_appendix}), this approach follows established practice, as is previously used by literature studying gender bias in embeddings or corpus~\cite{betti2023large,Caliskan_2017,charlesworth2021gender}.

\begin{table*}[h]

\centering
    \caption{List of attribute word sets used in the Word Embedding Association Test (WEAT), originally derived from~\cite{Caliskan_2017,betti2023large} and subsequently expanded by the authors to enrich semantic coverage and improve gender-related classification.} 
    \scalebox{0.9}{
    \begin{tabular}{p{0.2\linewidth}  p{0.8\linewidth}}
    
    \toprule
    \textbf{Attribute Set} & \textbf{Words} \\
    \midrule
\token{female} & \token{woman}, \token{women}, \token{she}, \token{mother}, \token{mothers}, \token{girl}, \token{girls}, \token{aunt}, \token{aunts},
\token{sister}, \token{sisters}, \token{queen}, \token{queens}, \token{lady}, \token{ladies}, \token{wife}, \token{wives},
\token{madam}, \token{madams}, \token{actress}, \token{actresses}, \token{niece}, \token{nieces}, \token{princess}, \token{princesses},
\token{braid}, \token{braids}, \token{bride}, \token{brides}, \token{mom}, \token{moms}, \token{grandmother}, \token{grandmothers},
\token{daughter}, \token{daughters}, \token{goddess}, \token{goddesses}, \token{nun}, \token{nuns}, \token{ladyship}, \token{ladyships},
\token{baroness}, \token{baronesses}, \token{empress}, \token{empresses}, \token{witch}, \token{witches}, \token{duchess}, \token{duchesses} \\
\midrule
\token{male} & \token{man}, \token{men}, \token{he}, \token{father}, \token{fathers}, \token{boy}, \token{boys}, \token{uncle}, \token{uncles},
\token{brother}, \token{brothers}, \token{king}, \token{kings}, \token{gentleman}, \token{gentlemen}, \token{husband}, \token{husbands},
\token{sir}, \token{sirs}, \token{actor}, \token{actors}, \token{nephew}, \token{nephews}, \token{prince}, \token{princes},
\token{beard}, \token{beards}, \token{groom}, \token{grooms}, \token{dad}, \token{dads}, \token{grandfather}, \token{grandfathers},
\token{son}, \token{sons}, \token{godfather}, \token{godfathers}, \token{monk}, \token{monks}, \token{lord}, \token{lords},
\token{baron}, \token{barons}, \token{emperor}, \token{emperors}, \token{knight}, \token{knights}, \token{wizard}, \token{wizards}, \token{duke}, \token{dukes} \\
    \bottomrule
    \end{tabular}}

    \label{tab:attribute_set_appendix} 
\end{table*}

\subsection{Details on contrastive loss function }
\label{sec:contras_loss_de}

The loss function is designed to jointly optimize the embedding space by drawing semantically similar word pairs closer together while pushing dissimilar pairs apart. For similar pairs, a linear pull term proportional to the predicted distance is introduced, promoting tighter semantic clustering. To ensure numerical stability and prevent excessively large gradients, we deliberately avoid logarithmic or high-degree penalty functions, opting instead for a linear formulation. For dissimilar pairs, we employ a standard contrastive loss component, which penalizes instances where the predicted distance falls below a predefined margin, thereby enforcing a minimum separation between unrelated terms.

While pre-training aims to minimize the $\ell_2$ distance between embeddings, the verification process instead focuses on finding the maximum verifiable radius under the $\ell_\infty$ norm, since formal verification over the $\ell_2$ domain is not directly possible using zonotopes. The authors intentionally adopt the $\ell_2$ distance for pre-training because it minimizes the overall distance between points, rather than just the maximum per-dimension difference. This encourages the model to learn more generalized and robust representations. Furthermore, the authors observed in their experiments that after pre-training, the $\ell_2$ distance between semantically similar words (such as gendered word pairs) is often slightly larger than their corresponding $\ell_\infty$  distance (as can be seen in \Cref{sec:distance_complete}). This supports the idea that utilizing the $\ell_2$ norm benefits downstream verification.

\subsection{Embedding model pre-training data construction}
\label{sec:data_cons_pre}
Given these limitations and the impracticality of verifying large-scale contextualized models like BERT~\cite{shi2020robustness,bonaert2021fast}, we pre-train a Keras embedding layer~\cite{ketkar2017introduction}.
 To construct an effective training set for contrastive loss optimization, we curated word pairs based on the semantic gender relationship using a heuristic approach. Each \token{male}-labeled word was explicitly paired with the terms \token{woman} and \token{female} to anchor the gender axis, and further matched with a diverse sample of 50-100 \token{female}-labeled words to form similar (label=0) pairs. This ensures that embeddings of gender-related terms are pulled closer in the vector space. To balance the contrastive objective, we generated dissimilar (label=1) pairs by selecting words from a broad English vocabulary that have no direct lexical or semantic relationship (e.g., synonyms, derivations) to any gender-specific term. This distinction was enforced by filtering out related terms using WordNet relations such as synonyms, hypernyms, and derivationally related forms. 

 To enhance the general semantic capacity of the embedding space, particularly because the embedding layer is later frozen during task-specific backpropagation, we conducted an auxiliary contrastive training phase focused on common semantic relationships such as synonymy (This general semantic training step proved essential, as model performance degraded significantly in its absence. This outcome is expected, given that the embedding layer, when integrated within the transformer-based architecture, is frozen during downstream training and therefore does not receive gradient updates to refine its semantic representations). Using WordNet, we generated word pairs labeled as similar (label=0) by extracting synonym pairs, while carefully excluding any gender-specific terms to preserve the integrity of gender representations learned in earlier pre-training stages, thereby maintaining the intended semantic distances between gendered terms. To form the dissimilar (label=1) pairs, we randomly sampled words from the general vocabulary that were not semantically related (each non-gendered word is paired with all valid WordNet synonyms, excluding any gendered terms, resulting in a variable number of synonym pairs per word. To ensure a balanced dataset, an equal number of dissimilar word pairs is generated, maintaining a one-to-one ratio with the total number of synonym pairs), again ensuring that no gender-specific tokens were involved. This allowed the model to learn to cluster semantically similar words and push apart unrelated ones. By training the embedding space on these general lexical relationships, we improve its ability to capture broader semantic structure, which is especially important since the embeddings are utilized as fixed features in downstream transformer-based language model text classification tasks. This step ensures that the embeddings are not only gender-aware but also semantically meaningful in a general linguistic context.

For pre-training the toxicity-related embedding model, a similar pre-training strategy was employed—using a contrastive learning framework to pull semantically similar words closer and push dissimilar words apart (to handle rare or profane words not explicitly found in the tokenizer vocabulary, we use BERT’s WordPiece-based subword tokenization. This strategy breaks out-of-vocabulary terms into smaller known units (e.g., \token{mother-fucker} becomes [\token{mother}, \token{-}, \token{fuck}, \token{\#\#er}]), allowing consistent representation throughout the embedding layer pre-training and language model training phases without full word-level coverage. These subwords are embedded and mean-pooled, enabling the model to capture semantic information even for obfuscated or stylized expressions). The strength of the pulling and pushing mechanisms (i.e., the values of $\alpha$ and the margin), along with hyperparameters, can be found in \Cref{sec:hyperparaneter_tuning}. The distance between gendered words, toxicity words, synonyms, and unrelated words before and after pre-training of embedding can be found in \Cref{sec:distance_complete}.

\subsection{Adult census dataset preprocessing}
\label{sec:data_cons_pre_tab}
Transformers require one-dimensional token sequences as input, whereas tabular data is inherently two-dimensional~\cite{badaro2023transformers}; however, the demographic information in the Adult Census dataset is structured in a tabular format, with each attribute represented as a separate column. To bridge this gap, the data and its contextual structure must be serialized before being processed by the model~\cite{badaro2023transformers}. A critical step in adapting tabular data for transformer-based models is converting the tabular structure into a linearized, one-dimensional format. Common strategies include:
column-wise concatenation of values, insertion of special separator tokens between fields~\cite{badaro2023transformers, 10.1145/3514221.3517906}, and use of text templates to represent each row as a structured sentence~\cite{chen2019tabfact,suadaa2021towards,chen2020logical}. More recent approaches move beyond predefined templates by generating natural language sentences directly from tabular data. This is achieved by fine-tuning sequence-to-sequence models such as T5~\cite{raffel2020exploring} or GPT~\cite{radford2019language}, as demonstrated in prior works~\cite{thorne2021database,neeraja2021incorporating}. We adopt simple concatenation for its straightforward implementation and effectiveness in our setting.

To adapt the Adult Census dataset for use with a sentence-to-label transformer-based binary classification model, we converted the structured tabular data into natural language sentences. The original dataset contains both numerical and categorical features, along with information indicating whether an individual's income exceeds \$50K. Since transformer models (e.g., BERT) operate on text sequences, we constructed a heuristic template that combines relevant features into coherent sentences. Specifically, attributes such as workclass, education, occupation, sex, marital status, relationship, race, native country, and hours per week were embedded into a natural language sentence format. The salary attribute, serving as the ground truth label for prediction, was intentionally excluded from the sentence construction. This transformation preserves the semantic content of the original features while enabling the use of sentence-based binary text classification models on structured tabular data. An example tabular data instance before and after sentence construction can be found in \Cref{tab:sample_instance_adult_tab_sen} (the feature "fnlwgt", along with several other complex numerical attributes, was excluded from this experiment, as the authors deliberately chose to focus on leveraging textual, simple numerical, and categorical features for model training and representation).

\begin{table*}
\centering
\small
\caption{Sample data instance from the Adult Census dataset, along with the constructed input sentence.}
\begin{tabular}{@{}lllll@{}}
\toprule
\rowcolor{gray!15}
\multicolumn{5}{l}{\textbf{Structured tabular data instance}} \\
\midrule
\textbf{Workclass} & \textbf{Education} & \textbf{Edu-Num} & \textbf{Occupation} & \textbf{Sex} \\
Private & 7th-8th grade & 4 & Machine-op-inspct & female \\
\midrule
\textbf{Marital Status} & \textbf{Relationship} & \textbf{Race} & \textbf{Country} & \textbf{Hours/Week} \\
Divorced & Unmarried & White & United-States & 40 \\
\midrule
\rowcolor{gray!15}
\multicolumn{5}{l}{\textbf{Constructed input sentence}} \\
\multicolumn{5}{@{}p{0.90\textwidth}@{}}{
\textit{A person's workclass is Private, education is 7th-8th grade (number of years of education is 4), occupation is Machine-op-inspct, sex is female, marital status is Divorced, relationship is Unmarried, race is White, native country is United-States, and they work 40 hours per week.}
} \\
\bottomrule
\end{tabular}

\label{tab:sample_instance_adult_tab_sen}
\end{table*}

\subsection{Data augmentation}

Text classification is a fundamental task in natural language processing, and achieving high performance often depends heavily on the size and quality of the training data~\cite{wei2019eda,kobayashi2018contextual}. Automatic data augmentation is widely employed in machine learning and computer vision, among other domains, to enhance model robustness, particularly when working with limited datasets~\cite{10.5555/645754.668381,szegedy2015going,10.1145/3065386,7113823,ko2015audio}.

A commonly used data augmentation strategy is synonym replacement, where selected words are substituted with their synonyms. This can be achieved using handcrafted lexical resources like WordNet~\cite{10.1145/219717.219748,zhang2016yelp}, word similarity computations~\cite{wang-yang-2015-thats}, or contextual embedding-based methods~\cite{kobayashi2018contextual}.

We employed a simple WordNet-based synonym replacement strategy to augment the dataset in the salary prediction experiments, aiming to increase data volume and enhance classification accuracy. This approach led to a notable improvement in test accuracy, rising from 80\% to around 90\% (we utilized straightforward WordNet synonym replacement, as empirical observations indicated that this method is already effective enough in improving model performance).

It is worth noting that data augmentation was performed while maintaining balanced class distributions across the dataset. An example of a sentence before and after applying synonym replacement prior to the training on the transformer model for the salary prediction task can be found in \Cref{tab:styled_sentence_comparison}, which was part of our investigation into gender fairness. The replacement probability was set to be 30\%.

\begin{table*}
\renewcommand{\arraystretch}{1.5}
\centering
\caption{Original vs. synonym-augmented sentence after data augmentation}
\begin{tabular}{p{0.9\linewidth}}
\toprule
\rowcolor[gray]{0.9} 
\textbf{Constructed input sentence} \\
\midrule
\textit{A person’s} workclass \textit{is} Self-emp-inc, \textit{sex is} male, \textit{education is} Bachelors \textit{(number of years of education is 13)}, \textit{occupation is} Exec-managerial, \textit{marital status is} Married-civ-spouse, \textit{relationship is} Husband, \textit{race is} White, \textit{native country is} United-States, \textit{and they work} 52 \textit{hours per week.} \\
\midrule
\rowcolor[gray]{0.9} 
\textbf{Augmented input sentence (synonym replacement)} \\
\midrule
\textit{A person’s} workclass \textit{is} Self-emp-inc, \textit{sex be} male, \textit{educational activity is} Bachelors \textit{(number of years of education be 13)}, \textit{occupation is} Exec-managerial, \textit{marital status is} Married-civ-spouse, \textit{human relationship is} Husband, \textit{race is} White, \textit{aborigine country is} United-States, \textit{and they work} 52 \textit{hours per week.} \\
\bottomrule
\end{tabular}

\label{tab:styled_sentence_comparison}
\end{table*}

\section{Evaluation details and ablation studies}

\subsection{Evaluation details} 
\label{sec:hyperparaneter_tuning}

It is important to note that all parameters—including the margin and $\alpha$ values used during embedding pre-training, as well as the hyperparameters for model training—must be carefully tuned to achieve both high accuracy and substantial verifiable radii. Comprehensive details on these hyperparameters are provided in the appendix. Specifically, the hyperparameters used for pre-training the Keras embedding layer are listed in \Cref{tab:pretraining_hyper}, while those utilized for training the language models are presented in \Cref{tab:transformer_hyperparameters_final}.    

The selection of margin and $\alpha$ values is based on empirical observation. The goal is not only to ensure that gendered words are embedded in close proximity but also to promote the clustering of synonyms while pushing antonyms farther apart, without compromising the relative closeness of gendered word pairs. The margin values are also selected empirically based on the nuanced relationship between the embedding space and the resulting verifiable radius. A detailed study of this relationship is provided in \Cref{sec:embandraduis}. The loss function used in this study is focal loss~\cite{Lin_2017_ICCV}, a training-time calibration technique that directly integrates class imbalance handling into the learning process. It works by adapting the loss function to emphasize challenging, misclassified samples.

To evaluate the fairness score of the models, we use the MATLAB toolbox CORA~\cite{althoff2015introduction}, 
which enables formal verification using set-based computing~\cite{kochdumper2022open,ladner2023automatic,bonaert2021fast}.
Please visit \Cref{sec:zonotope} for a detailed explanation of this process using zonotopes.

\begin{table*}
\centering
\caption{Parameters used during embedding model pre-training (Gen. Sem. Train. refers to general semantic pre-training, such as synonym clustering).}
\begin{tabularx}{\linewidth}{l *{6}{>{\centering\arraybackslash}X}}
\toprule
 & \multicolumn{6}{c}{\cellcolor{gray!6}\textbf{Salary Prediction \& Gender Bias Mitigation}} \\
\cmidrule(lr){2-7}
\textbf{Model} & \boldmath$\alpha$ & \textbf{Margin} & \boldmath$\alpha$ (Gen. Sem. Train.) & \textbf{Margin} (Gen. Sem. Train.) & \textbf{Epoch} & \textbf{Epoch} (Gen. Sem. Train.) \\
\midrule
6 blocks & 1000 & 1.0 & 1.0 & 0.5 & 150 & 30 \\
8 blocks & 1000 & 1.0 & 1.0 & 0.8 & 150 & 30 \\
10 blocks & 1000 & 1.0 & 1.0 & 0.8 & 150 & 30 \\
\midrule
 & \multicolumn{6}{c}{\cellcolor{gray!10}\textbf{Toxicity Detection \& Content Moderation}} \\
\cmidrule(lr){2-7}
\textbf{Model} & \boldmath$\alpha$ & \textbf{Margin} & \boldmath$\alpha$ (Gen. Sem. Train.) & \textbf{Margin} (Gen. Sem. Train.) & \textbf{Epoch} & \textbf{Epoch} (Gen. Sem. Train.) \\
\midrule
3 blocks & 20 & 0.5 & 1.0 & 0.3 & 100 & 100 \\
\bottomrule
\end{tabularx}

\label{tab:pretraining_hyper}
\end{table*}

\begin{table*}
\centering

\setlength{\tabcolsep}{3.0mm}  
\caption{Language model training hyperparameters used for salary prediction and toxicity detection tasks. 
The table includes vocabulary size, maximum sequence length, embedding dimension, number of attention heads, feedforward network dimension, dropout rate, number of training epochs, batch size, and optimization strategy.}
\begin{tabular}{lcccccccc}
\toprule
 & \multicolumn{8}{c}{\cellcolor{gray!10}\textbf{Salary Prediction \& Gender Bias Mitigation}} \\
\cmidrule(lr){2-9}
\textbf{Model} & Vocab (BERT) & Seq. Len. & Embed Dim & Heads & FF Dim & Dropout & Epochs & Batch \\
\midrule
6 blocks & 30,522 & 200 & 8 & 2 & 8 & 0.2 & 100 & 32 \\
8 blocks & 30,522 & 200 & 8 & 2 & 8 & 0.1 & 100 & 32 \\
10 blocks & 30,522 & 200 & 8 & 2 & 8 & 0.01 & 100 & 32 \\
\midrule
 & \multicolumn{8}{c}{\cellcolor{gray!10}\textbf{Toxicity Detection \& Content Moderation}} \\
\cmidrule(lr){2-9}
\textbf{Model} & Vocab & Seq. Len. & Embed Dim & Heads & FF Dim & Dropout & Epochs & Batch \\
\midrule
3 blocks & 30,522 & 200 & 8 & 2 & 8 & 0.2 & 150 & 128 \\
\midrule
\multicolumn{9}{l}{\cellcolor{gray!20}\textbf{Optimization \& training strategy}} \\
\multicolumn{9}{l}{\textbf{Optimizer:} AdamW (learning rate = 1e-3, weight decay = 1e-5)} \\
\multicolumn{9}{l}{\textbf{LR scheduler:} ReduceLROnPlateau (monitor = val\_loss, factor = 0.5, patience = 3, min\_lr = 1e-6)} \\
\multicolumn{9}{l}{\textbf{Early stopping:} Patience = 10, restore\_best\_weights = True} \\
\multicolumn{9}{l}{\textbf{Loss function:} Focal Loss ($\alpha$ = 0.25, $\gamma$ = 2.0)} \\
\bottomrule
\end{tabular}

\label{tab:transformer_hyperparameters_final}
\end{table*}

\subsection{Sentence-wise maximum verifiable perturbation radius}
 \label{sec:maxraduis}

\begin{table*}
\centering
\caption{Embedding bounds and maximum verifiable noise radii for the salary prediction model (8 blocks).}
\begin{tabular}{l c c c c}
\toprule
\textbf{Model} &
\multicolumn{4}{c}{\cellcolor{gray!10}\textbf{Salary Prediction \& Gender Bias Mitigation (8 blocks)}} \\
\midrule
\multirow{3}{*}{\textbf{Embedding Bounds}} 
 & \multicolumn{2}{r}{Minimum ($\inf$)} & \multicolumn{2}{l}{-1.696724} \\
 & \multicolumn{2}{r}{Maximum ($\sup$)} & \multicolumn{2}{l}{1.735214} \\
 & \multicolumn{2}{r}{Range ($\sup - \inf$)} & \multicolumn{2}{l}{3.431938} \\

\cmidrule(lr){2-5}
 & \multicolumn{2}{c}{\textbf{Sentences 1--25}}
 & \multicolumn{2}{c}{\textbf{Sentences 26--50}} \\
\cmidrule(lr){2-3} \cmidrule(lr){4-5}
 & \textbf{Sentence}
 & \makecell{\textbf{Max verifiable}\\\textbf{noise radius} ($\uparrow$)}
 & \textbf{Sentence}
 & \makecell{\textbf{Max verifiable}\\\textbf{noise radius} ($\uparrow$)} \\
\midrule

\multirow{25}{*}{\textbf{Noise Radii}}
 & Sentence 1  & 8.6079e-04 & Sentence 26 & 8.2806e-04 \\
 & Sentence 2  & 4.9749e-04 & Sentence 27 & 7.7405e-04 \\
 & Sentence 3  & 2.9457e-04 & Sentence 28 & 8.5915e-04 \\
 & Sentence 4  & 8.8533e-04 & Sentence 29 & 8.5915e-04 \\
 & Sentence 5  & 6.5950e-04 & Sentence 30 & 8.7715e-04 \\
 & Sentence 6  & 8.5915e-04 & Sentence 31 & 8.5915e-04 \\
 & Sentence 7  & 6.5459e-04 & Sentence 32 & 3.1748e-04 \\
 & Sentence 8  & 6.8896e-04 & Sentence 33 & 9.3279e-04 \\
 & Sentence 9  & 7.2169e-04 & Sentence 34 & 6.8896e-04 \\
 & Sentence 10 & 2.8147e-04 & Sentence 35 & 5.9895e-04 \\
 & Sentence 11 & 6.8896e-04 & Sentence 36 & 7.8060e-04 \\
 & Sentence 12 & 8.5915e-04 & Sentence 37 & 7.2823e-04 \\
 & Sentence 13 & 8.5915e-04 & Sentence 38 & 2.9457e-04 \\
 & Sentence 14 & 8.5751e-04 & Sentence 39 & 3.5512e-04 \\
 & Sentence 15 & 6.8896e-04 & Sentence 40 & 8.2806e-04 \\
 & Sentence 16 & 9.4752e-04 & Sentence 41 & 8.2806e-04 \\
 & Sentence 17 & 4.7949e-04 & Sentence 42 & 8.1987e-04 \\
 & Sentence 18 & 8.2806e-04 & Sentence 43 & 8.2806e-04 \\
 & Sentence 19 & 5.8586e-04 & Sentence 44 & 7.9696e-04 \\
 & Sentence 20 & 8.0515e-04 & Sentence 45 & 1.9638e-04 \\
 & Sentence 21 & 2.9457e-04 & Sentence 46 & 9.1152e-04 \\
 & Sentence 22 & 6.1368e-04 & Sentence 47 & 8.3624e-04 \\
 & Sentence 23 & 6.7586e-04 & Sentence 48 & 8.5915e-04 \\
 & Sentence 24 & 8.5915e-04 & Sentence 49 & 8.2806e-04 \\
 & Sentence 25 & 6.9878e-04 & Sentence 50 & 7.6751e-04 \\
\bottomrule
\end{tabular}

\label{tab:embedding_bounds_salary}
\end{table*}

\begin{table*}
\centering

\caption{Embedding bounds and maximum verifiable noise radii for the toxicity detection model (3 blocks).}
\begin{tabular}{l c c c c}
\toprule
\textbf{Model} &
\multicolumn{4}{c}{\cellcolor{gray!10}\textbf{Toxicity Detection \& Content Moderation (3 blocks)}} \\
\midrule
\multirow{3}{*}{\textbf{Embedding Bounds}} 
 & \multicolumn{2}{r}{Minimum ($\inf$)} & \multicolumn{2}{l}{-0.838544} \\
 & \multicolumn{2}{r}{Maximum ($\sup$)} & \multicolumn{2}{l}{0.865149} \\
 & \multicolumn{2}{r}{Range ($\sup - \inf$)} & \multicolumn{2}{l}{1.703692} \\

\cmidrule(lr){2-5}
 & \multicolumn{2}{c}{\textbf{Sentences 1--25}}
 & \multicolumn{2}{c}{\textbf{Sentences 26--50}} \\
\cmidrule(lr){2-3} \cmidrule(lr){4-5}
 & \textbf{Sentence}
 & \makecell{\textbf{Max verifiable}\\\textbf{noise radius} ($\uparrow$)}
 & \textbf{Sentence}
 & \makecell{\textbf{Max verifiable}\\\textbf{noise radius} ($\uparrow$)} \\
\midrule

\multirow{25}{*}{\textbf{Noise Radii}}
 & Sentence 1  & 4.2090e-03 & Sentence 26 & 4.3820e-03 \\
 & Sentence 2  & 6.3423e-03 & Sentence 27 & 4.9913e-03  \\
 & Sentence 3  & 4.6858e-03 & Sentence 28 & 6.2285e-03  \\
 & Sentence 4  & 4.0839e-03 & Sentence 29 & 4.0562e-03  \\
 & Sentence 5  & 4.4446e-03 & Sentence 30 & 3.5509e-03  \\
 & Sentence 6  & 4.8865e-03 & Sentence 31 & 4.3999e-03  \\
 & Sentence 7  & 4.8061e-03 & Sentence 32 & 3.3641e-03  \\
 & Sentence 8  & 5.5104e-03 & Sentence 33 & 3.0164e-03 \\
 & Sentence 9  & 3.8215e-03 & Sentence 34 & 3.0164e-03  \\
 & Sentence 10 & 4.6940e-03 & Sentence 35 & 4.5656e-03  \\
 & Sentence 11 & 3.7451e-04 & Sentence 36 & 3.3454e-03  \\
 & Sentence 12 & 5.2959e-03 & Sentence 37 & 5.4438e-03  \\
 & Sentence 13 & 4.6745e-03 & Sentence 38 & 4.0497e-03  \\
 & Sentence 14 & 1.4160e-03 & Sentence 39 & 5.1343e-03  \\
 & Sentence 15 & 5.0148e-03 & Sentence 40 & 3.3015e-03 \\
 & Sentence 16 & 4.7354e-03 & Sentence 41 & 4.4242e-03 \\
 & Sentence 17 & 5.2748e-03 & Sentence 42 & 3.9230e-03  \\
 & Sentence 18 & 4.7403e-03 & Sentence 43 & 4.0457e-03  \\
 & Sentence 19 & 5.7736e-03 & Sentence 44 & 7.1896e-04  \\
 & Sentence 20 & 2.9083e-03 & Sentence 45 & 4.3341e-03  \\
 & Sentence 21 & 2.8401e-03 & Sentence 46 & 3.1326e-03  \\
 & Sentence 22 & 5.3471e-03 & Sentence 47 & 4.9742e-03  \\
 & Sentence 23 & 5.0067e-03 & Sentence 48 & 4.8288e-03  \\
 & Sentence 24 & 4.7801e-03 & Sentence 49 & 4.5640e-03  \\
 & Sentence 25 & 3.5371e-03 & Sentence 50 & 5.7557e-03  \\
\bottomrule
\end{tabular}

\label{tab:embedding_bounds_toxicity}
\end{table*}

The maximum verifiable perturbed radius is reported in this section (see \Cref{tab:embedding_bounds_salary,tab:embedding_bounds_toxicity}) across a set of 20 randomly selected sentences, each comprising 20 tokens. The sample includes an equal distribution of 10 sentences from each class label, drawn from the training dataset. Example per-sentence radius values for a representative model from each experiment are provided in this Appendix\footnote{Note that the presence of words outside the verifiable radius does not necessarily indicate a change in the model’s classification output. Given that the verification method used in this study is incomplete, such instances may still result in correct classifications.}.

\subsection{Ablation studies}
\label{sec:ablation-study}

\subsubsection{Adversarial Attack}

Robustness is a key prerequisite for reliable model verification, and adversarial training makes the models more robust \cite{goodfellow2015explainingharnessingadversarialexamples,ijcai2021p591}; therefore, in this section, we examine whether our framework assigns higher fairness scores to models that undergo adversarial training. To better show the effect of adversarial training, we consider a model variant with embedding layers that differ from those used in \Cref{sec:result_main}. This design choice ensures that the conventionally trained model exhibits a fairness score of zero, providing a clear baseline model from which improvements introduced by adversarial training can be more clearly measured. Please note that the 2 models used for this ablation study have the exact same embedding layers so that we can isolate the effect of pretraining.

To improve robustness against perturbations, we apply \emph{offline adversarial data augmentation} via synonym replacement. Given an existing labeled dataset, additional training samples are generated \emph{only} for the positive class ($y = 1$), while all original samples are retained. The number of variants per positive instance is controlled by an \emph{augmentation factor} $k$. In our adversarial experiments, we set the augment factor to be 3, producing three additional augmented sentences for each original positive example. Detailed pretraining and training hyperparameters can be found in \Cref{tab:combined_results_adv}.

Each augmented sentence is created through token-level synonym substitution based on WordNet. For every non-numeric token in a sentence, synonym replacement is applied independently with probability $p$. We use a high replacement probability of $p = 0.8$, which encourages the generation of more adversarial examples. This augmentation strategy introduces perturbations, thereby simulating adversarial paraphrases that increase the model’s robustness during training. The result can be found in \Cref{tab:combined_results_adv}.

\begin{table*}[t]
\centering
\small
\setlength{\tabcolsep}{4pt}
\renewcommand{\arraystretch}{0.9}
\caption{Training configuration, optimization strategy, and summary results for baseline and adversarially trained models. Fairness improvements are achieved through offline synonym-based augmentation during model training.}
\begin{tabularx}{\linewidth}{l *{8}{>{\centering\arraybackslash}X}}
\toprule

 &
\multicolumn{8}{c}{\cellcolor{gray!6}\textbf{Pretraining Hyperparameters and Augmentation Parameters}} \\
\cmidrule(lr){2-9}

\textbf{Model} & $\boldsymbol{\alpha}$
& Margin
& $\boldsymbol{\alpha}$ (Gen. Sem.)
& Margin (Gen. Sem.)
& Epochs
& Epochs (Gen. Sem.)
& Aug. Prob.
& Aug. Factor \\

\midrule
3 blocks (baseline)
& 20 & 0.3 & 1.0 & 0.3 & 100 & 100 & 0 & 0 \\

3 blocks (adv. trained)
& 20 & 0.3 & 1.0 & 0.3 & 100 & 100 & 0.8 & 3 \\

\midrule

\multicolumn{9}{c}{\cellcolor{gray!10}\textbf{Toxicity Detection \& Content Moderation Model Configuration}} \\
\midrule

\textbf{Model}
& Vocab
& Seq. Len.
& Embed Dim
& Heads
& FF Dim
& Dropout
& Epochs
& Batch \\

\midrule
3 blocks
& 30{,}522
& 200
& 8
& 2
& 8
& 0.2
& 150
& 128 \\

\midrule

\multicolumn{9}{l}{\cellcolor{gray!20}\textbf{Optimization \& Training Strategy}} \\
\multicolumn{9}{l}{\textbf{Optimizer:} AdamW (lr $=10^{-3}$, weight decay $=10^{-5}$)} \\
\multicolumn{9}{l}{\textbf{LR scheduler:} ReduceLROnPlateau (factor $=0.5$, patience $=3$, min lr $=10^{-6}$)} \\
\multicolumn{9}{l}{\textbf{Early stopping:} Patience $=10$, restore best weights} \\
\multicolumn{9}{l}{\textbf{Loss:} Focal Loss ($\alpha = 0.25$, $\gamma = 2.0$)} \\

\midrule

\multicolumn{5}{c}{\cellcolor{gray!15}\textbf{Evaluation Summary}} &
\multicolumn{4}{c}{\cellcolor{gray!15}\textbf{Embedding Bounds}} \\
\midrule

\multicolumn{6}{l}{
\begin{tabular}[t]{@{}l c c@{}}
\textbf{Model} & \textbf{Fairness Score} & \textbf{Test Acc.} \\
Baseline& 0\% & 91\% \\
Adversarially Trained& 44\% & 91\% 
\end{tabular}
}
&
\multicolumn{3}{l}{
\begin{tabular}[t]{@{}l c@{}}
Minimum ($\inf$) & -1.242537 \\
Maximum ($\sup$) & 1.161774 \\
Range ($\sup - \inf$) & 2.404311
\end{tabular}
} \\

\bottomrule
\end{tabularx}

\label{tab:combined_results_adv}
\end{table*}

\subsubsection{Fairness scores without pre-training the Keras embedding layer}

This section presents the results of the fairness assessment without explicitly pre-training the embedding model. It is important to note that the Keras embedding layer was unfrozen during training, allowing it to learn basic semantic representations of words solely through backpropagation. This fairness assessment is limited to gender bias and is intended solely for illustrative purposes—to demonstrate that, without pre-training, the model fails to exhibit verifiably fair behavior.

The results for the 50 nearest neighbors and their corresponding distances for the words \token{female} and \token{male} are presented in \Cref{tab:combined_8b_gender}, respectively. As shown, in the absence of pre-training, the nearest neighbors seldom reflect strongly gendered semantics. This underscores the importance of embedding layer pre-training in capturing meaningful semantic relationships, thereby justifying the design of our experimental framework.
For comparison, the distance results with pre-training are provided in \Cref{sec:distance_complete}, some examples of distances of strongly gender-related word pairs without specifically pre-training can be found in \Cref{tab:gender_inf_nopre}.

\begin{figure}[t]
\centering
\includegraphics[width=0.7\linewidth]{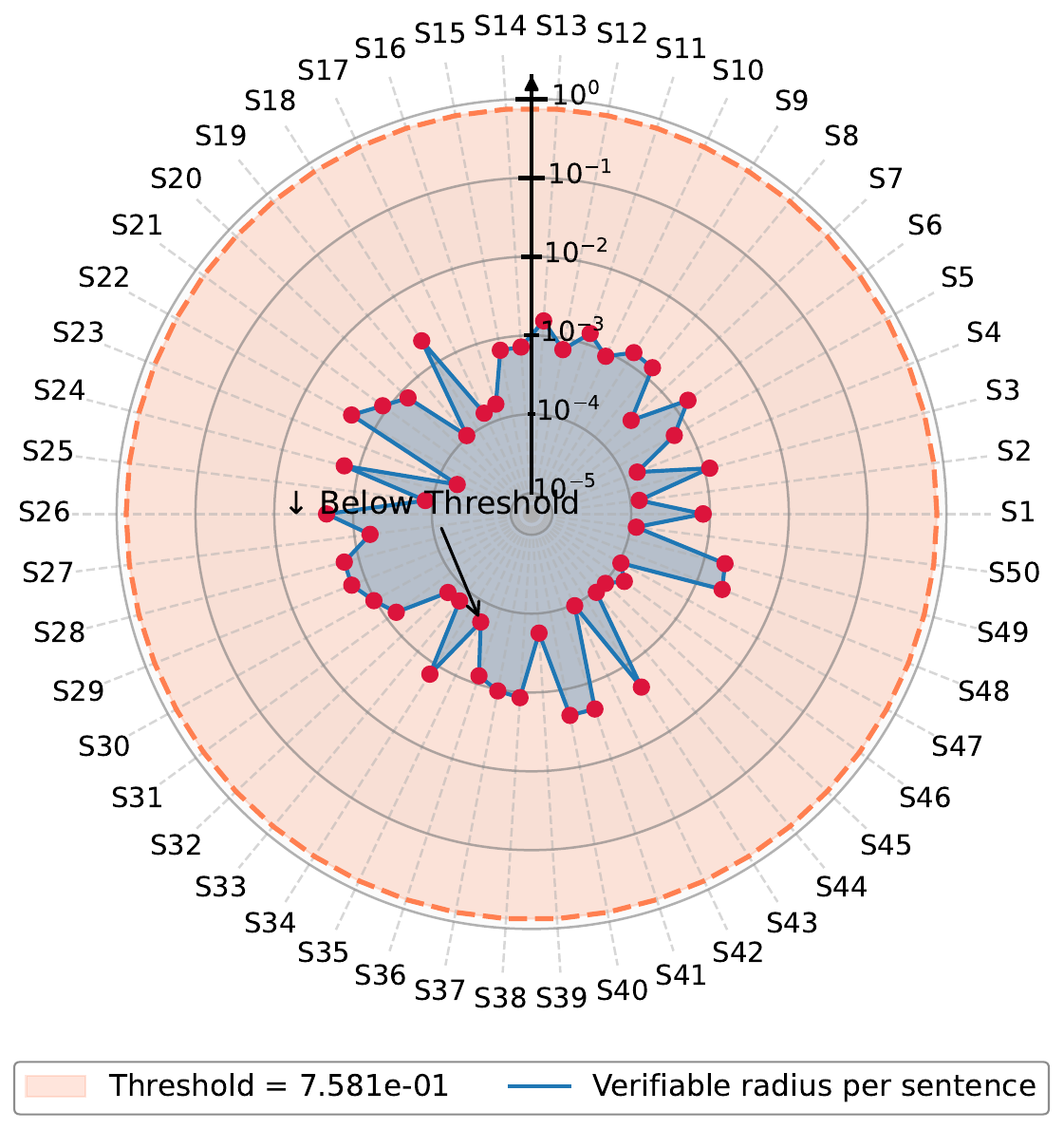}
\caption{Verifiability radar plot for the 8-block salary prediction model. Each axis corresponds to a sentence, with the blue region indicating its $\ell_\infty$-norm verifiable radius. The coral dashed circle denotes the safety threshold (0.7581), representing the $\ell_\infty$ distance between the tokens \token{female} and \token{male} in the Keras embedding layer without domain-specific pre-training.}
\label{fig:gender_8b_nopre_rader}
\end{figure}

For the sake of completeness, we conducted the fairness score assessment despite the absence of gendered words among the 50 nearest neighbors for both gender indicators, we simplified the fairness evaluation by setting the threshold based solely on the $\ell_\infty$  distance between \token{female} and \token{male} (which is the gender indicator in the Adult Census dataset), as the reference circle (which is 0.7581), rather than using the maximum distance among the 50 neighboring words of the two gender indicators. Please note that the entire experimental setup, including the language model training parameters, remains identical to the experiment that includes the pre-training of the embedding model (\text{Section 4.2 Experimental Results}). The results of the formal verification for the model without any pre-training of the Keras embedding layer are presented in \Cref{fig:gender_8b_nopre_rader}. As shown, not only do all sentences fail verification, but their corresponding $\ell_\infty$ radii also lie significantly within the reference circle.

\begin{table*}
\centering
\small
\setlength{\tabcolsep}{3pt}
\renewcommand{\arraystretch}{1.0}
\caption{Top 50 most $\ell_\infty$ -similar words to \token{female} and \token{male} using the Keras embedding layer \textbf{without} pre-training.}
\resizebox{\textwidth}{!}{%
\begin{tabular}{lllllllll}
\toprule
\textbf{Model} & \multicolumn{8}{c}{\cellcolor{gray!10}\textbf{Salary Prediction \& Gender Bias Mitigation (8 blocks)}} \\
\midrule
\textbf{} & \multicolumn{4}{c}{\textbf{\token{female}}} & \multicolumn{4}{c}{\textbf{\token{male}}} \\
\cmidrule(lr){2-5} \cmidrule(lr){6-9}
 & \textbf{Word} & \textbf{Distance} & \textbf{Word} & \textbf{Distance} & \textbf{Word} & \textbf{Distance} & \textbf{Word} & \textbf{Distance} \\
\midrule
\multirow{25}{*}{} 
 & \token{service}     & $6.710\text{e}{-02}$ & \token{handler}     & $8.93\text{e}{-02}$ & \token{husband}     & $9.55\text{e}{-02}$ & \token{bachelor}    & $1.133\text{e}{-01}$ \\
 & \token{fishing}     & $1.273\text{e}{-01}$ & \token{farming}     & $1.318\text{e}{-01}$ & \token{support}     & $1.190\text{e}{-01}$ & \token{widowed}     & $1.262\text{e}{-01}$ \\
 & \token{unmarried}   & $1.335\text{e}{-01}$ & \token{other}       & $1.421\text{e}{-01}$ & \token{inc}         & $1.367\text{e}{-01}$ & \token{protective}  & $1.395\text{e}{-01}$ \\
 & \token{10th}        & $1.427\text{e}{-01}$ & \token{2}                    & $1.450\text{e}{-01}$ & \token{75}                   & $1.420\text{e}{-01}$ & \token{france}      & $1.434\text{e}{-01}$ \\
 & \token{7th}         & $1.488\text{e}{-01}$ & \token{6th}         & $1.504\text{e}{-01}$ & \token{\#\#ec}      & $1.465\text{e}{-01}$ & \token{specialty}   & $1.474\text{e}{-01}$ \\
 & \token{5th}         & $1.560\text{e}{-01}$ & \token{8}                    & $1.672\text{e}{-01}$ & \token{85}                   & $1.477\text{e}{-01}$ & \token{\#\#em}      & $1.519\text{e}{-01}$ \\
 & \token{1st}         & $1.753\text{e}{-01}$ & \token{5}                    & $1.794\text{e}{-01}$ & \token{13}                   & $1.534\text{e}{-01}$ & \token{\#\#v}       & $1.568\text{e}{-01}$ \\
 & \token{armed}       & $1.831\text{e}{-01}$ & \token{absent}      & $1.841\text{e}{-01}$ & \token{45}                   & $1.605\text{e}{-01}$ & \token{16}                   & $1.713\text{e}{-01}$ \\
 & \token{8th}         & $1.851\text{e}{-01}$ & \token{6}                    & $1.954\text{e}{-01}$ & \token{self}        & $1.750\text{e}{-01}$ & \token{s}           & $1.801\text{e}{-01}$ \\
 & \token{relative}    & $1.973\text{e}{-01}$ & \token{7}                    & $1.978\text{e}{-01}$ & \token{ci}          & $1.785\text{e}{-01}$ & \token{55}                   & $1.850\text{e}{-01}$ \\
 & \token{own}         & $1.960\text{e}{-01}$ & \token{honduras}    & $2.079\text{e}{-01}$ & \token{60}                   & $1.859\text{e}{-01}$ & \token{52}                   & $1.862\text{e}{-01}$ \\
 & \token{cleaner}     & $2.103\text{e}{-01}$ & \token{12th}        & $2.138\text{e}{-01}$ & \token{italy}       & $1.894\text{e}{-01}$ & \token{90}                   & $1.911\text{e}{-01}$ \\
 & \token{4th}         & $2.151\text{e}{-01}$ & \token{scotland}    & $2.166\text{e}{-01}$ & \token{42}                   & $1.926\text{e}{-01}$ & \token{ninth}       & $1.941\text{e}{-01}$ \\
 & \token{salvador}    & $2.187\text{e}{-01}$ & \token{portugal}    & $2.214\text{e}{-01}$ & \token{ex}          & $1.965\text{e}{-01}$ & \token{spouse}      & $1.971\text{e}{-01}$ \\
 & \token{3}                    & $2.232\text{e}{-01}$ & \token{59}                   & $2.253\text{e}{-01}$ & \token{twelfth}     & $1.979\text{e}{-01}$ & \token{48}                   & $2.007\text{e}{-01}$ \\
 & \token{32}                   & $2.295\text{e}{-01}$ & \token{forces}      & $2.344\text{e}{-01}$ & \token{46}                   & $2.012\text{e}{-01}$ & \token{tech}        & $2.051\text{e}{-01}$ \\
 & \token{dominican}   & $2.369\text{e}{-01}$ & \token{\#\#vi}       & $2.384\text{e}{-01}$ & \token{england}     & $2.080\text{e}{-01}$ & \token{lee}         & $2.130\text{e}{-01}$ \\
 & \token{86}                   & $2.419\text{e}{-01}$ & \token{thailand}    & $2.427\text{e}{-01}$ & \token{39}                   & $2.139\text{e}{-01}$ & \token{prof}        & $2.148\text{e}{-01}$ \\
 & \token{37}                   & $2.432\text{e}{-01}$ & \token{11th}        & $2.497\text{e}{-01}$ & \token{15}                   & $2.171\text{e}{-01}$ & \token{degree}      & $2.182\text{e}{-01}$ \\
 & \token{married}     & $2.184\text{e}{-01}$ & \token{35}                   & $2.522\text{e}{-01}$ & \token{pac}         & $2.238\text{e}{-01}$ & \token{89}                   & $2.268\text{e}{-01}$ \\
 & \token{74}                   & $2.537\text{e}{-01}$ & \token{never}       & $2.587\text{e}{-01}$ & \token{edgar}       & $2.269\text{e}{-01}$ & \token{44}                   & $2.271\text{e}{-01}$ \\
 & \token{24}                   & $2.653\text{e}{-01}$ & \token{jamaica}     & $2.671\text{e}{-01}$ & \token{iran}        & $2.271\text{e}{-01}$ & \token{el}          & $2.277\text{e}{-01}$ \\
 & \token{peru}        & $2.674\text{e}{-01}$ & \token{4}                    & $2.675\text{e}{-01}$ & \token{nation}      & $2.286\text{e}{-01}$ & \token{ireland}     & $2.295\text{e}{-01}$ \\
 & \token{34}                   & $2.694\text{e}{-01}$ & \token{puerto}      & $2.750\text{e}{-01}$ & \token{65}                   & $2.297\text{e}{-01}$ & \token{43}                   & $2.305\text{e}{-01}$ \\
 & \token{92}                   & $2.756\text{e}{-01}$ & \token{china}       & $2.756\text{e}{-01}$ & \token{federal}     & $2.306\text{e}{-01}$ & \token{63}                   & $2.782\text{e}{-01}$ \\

\bottomrule
\end{tabular}}

\label{tab:combined_8b_gender}
\end{table*}

\begin{table}[t]
\centering
\caption{$\ell_\infty$ distances between gender-related word pairs without specifically pre-training the Keras embedding layer.}
\begin{tabular}{lclc}
\toprule
&\textbf{Pair} & & $\boldsymbol{\ell_\infty}$ \textbf{Distance} \\
\midrule
\token{female} & $\leftrightarrow$ & \token{male} & $7.581\text{e}{-01}$ \\
\token{female} & $\leftrightarrow$ & \token{man}  & $9.792\text{e}{-01}$ \\
\token{woman}  & $\leftrightarrow$ & \token{male} & $2.582\text{e}{-01}$ \\
\token{woman}  & $\leftrightarrow$ & \token{man}  & $6.116\text{e}{-01}$ \\
\token{he}     & $\leftrightarrow$ & \token{she}  & $8.770\text{e}{-02}$ \\
\bottomrule
\end{tabular}

\label{tab:gender_inf_nopre}
\end{table}

\subsubsection{Relationship between embedding space and max perturbed noise radius}
\label{sec:embandraduis}
\begin{figure*}
    \centering
    \includegraphics[width=0.75\linewidth]{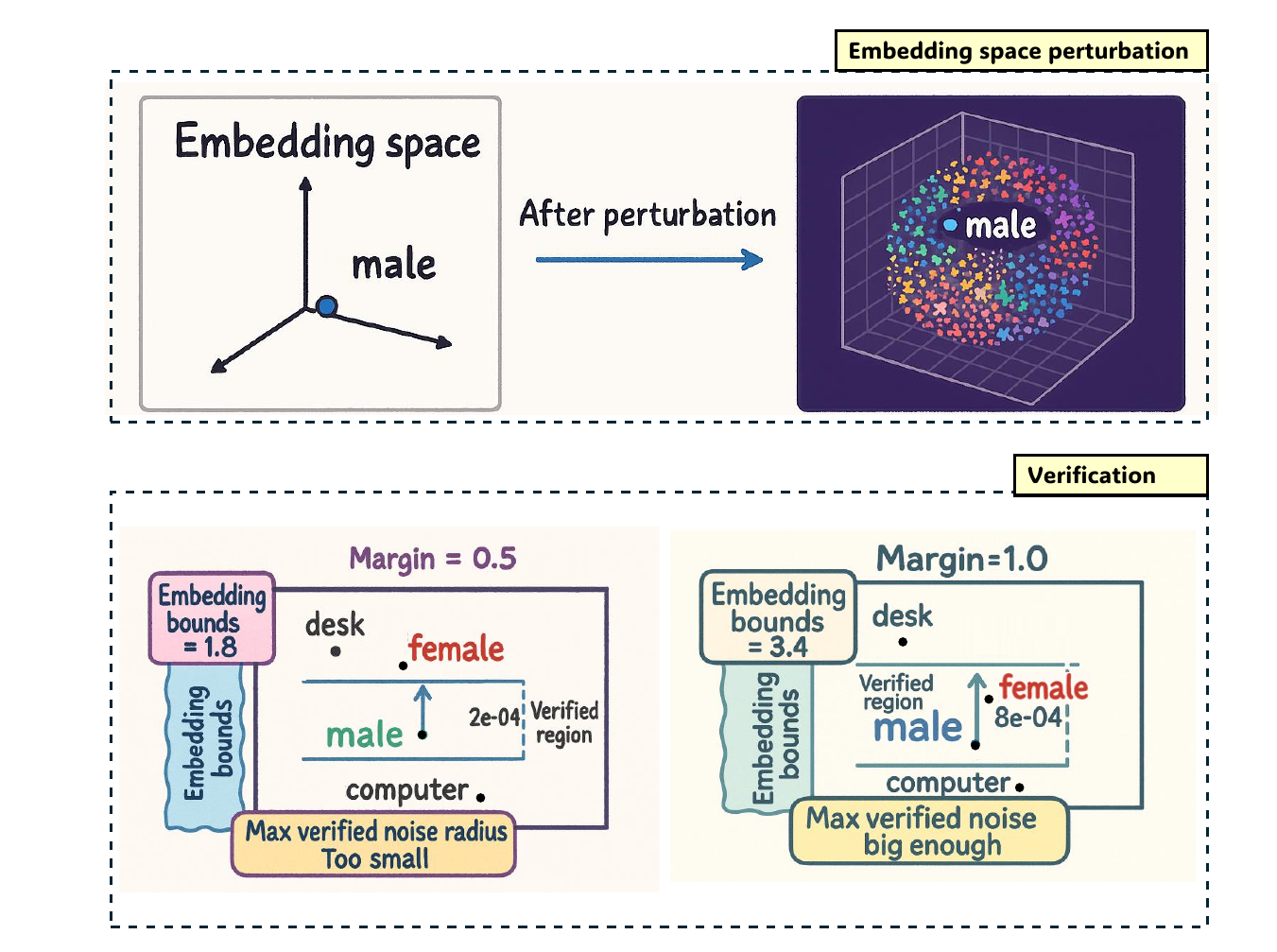}
    \caption{The effect of contrastive loss margin on embedding geometry and zonotope-based verification. A larger margin (right) results in more spread-out embeddings and enables larger certified perturbation bounds, while a smaller margin (left) leads to tighter clustering and limited verification robustness.}
    \label{fig:marg_emd}
\end{figure*}

This section investigates the relationship between the margin parameter in \eqref{eq:enhanced_contrastive_loss}, the resulting embedding bounds, and the maximum verifiable perturbation radius. As the margin parameter decreases, the embedding bounds contract accordingly, leading to a reduction in the verifiable radius. This contraction can impede the successful verification of semantically similar words, as their embedding distances may fall outside the increasingly tightened bounds. An illustrative graph \Cref{fig:marg_emd} highlights the trade-off between margin size and verifiability within semantic clusters (The range in the graph is for demonstrative purposes only; for more detailed metrics corresponding to specific margin values, please refer to \Cref{sec:maxraduis}).

As shown in the graph, when the margin is too small, meaning the repelling (pushing) force between unrelated terms is weak, it can result in a tighter embedding bound. The distances between related words (e.g., gendered word pairs) determined during the pre-training phase remain relatively small and stable. This behavior is consistent with the mathematical formulation in \eqref{eq:enhanced_contrastive_loss}, where the attractive force between related terms and the repulsive force between unrelated terms are combined through simple addition. Because they are added together rather than being interdependent through more complex interactions, these components remain mathematically separable. Consequently, an overly compact embedding space may cause related terms to fall outside the verifiable radius, as their distances remain generally unchanged after pre-training.

However, it has been empirically observed that the margin may have a slight influence on the final proximity of related word embeddings after pre-training, which could be attributed to the general property of the embedding space geometrically, but this effect is generally minimal and, in many cases, negligible. The authors further observed through empirical analysis that simply increasing the contrastive loss margin does not necessarily enhance the verification process. In fact, when the margin becomes excessively large, particularly in conjunction with an expanded embedding space, the verifiable noise radius does not always increase correspondingly. In some cases, it may even diminish. These findings highlight that the choice of margin is not a matter of maximizing separation or pulling force ($\alpha$), but rather one of carefully balancing semantic dispersion with geometric certifiability.

\subsubsection{Transformer-neighbor configurations}

\begin{figure*}
    \centering
    
    \includegraphics[width=0.75\linewidth]{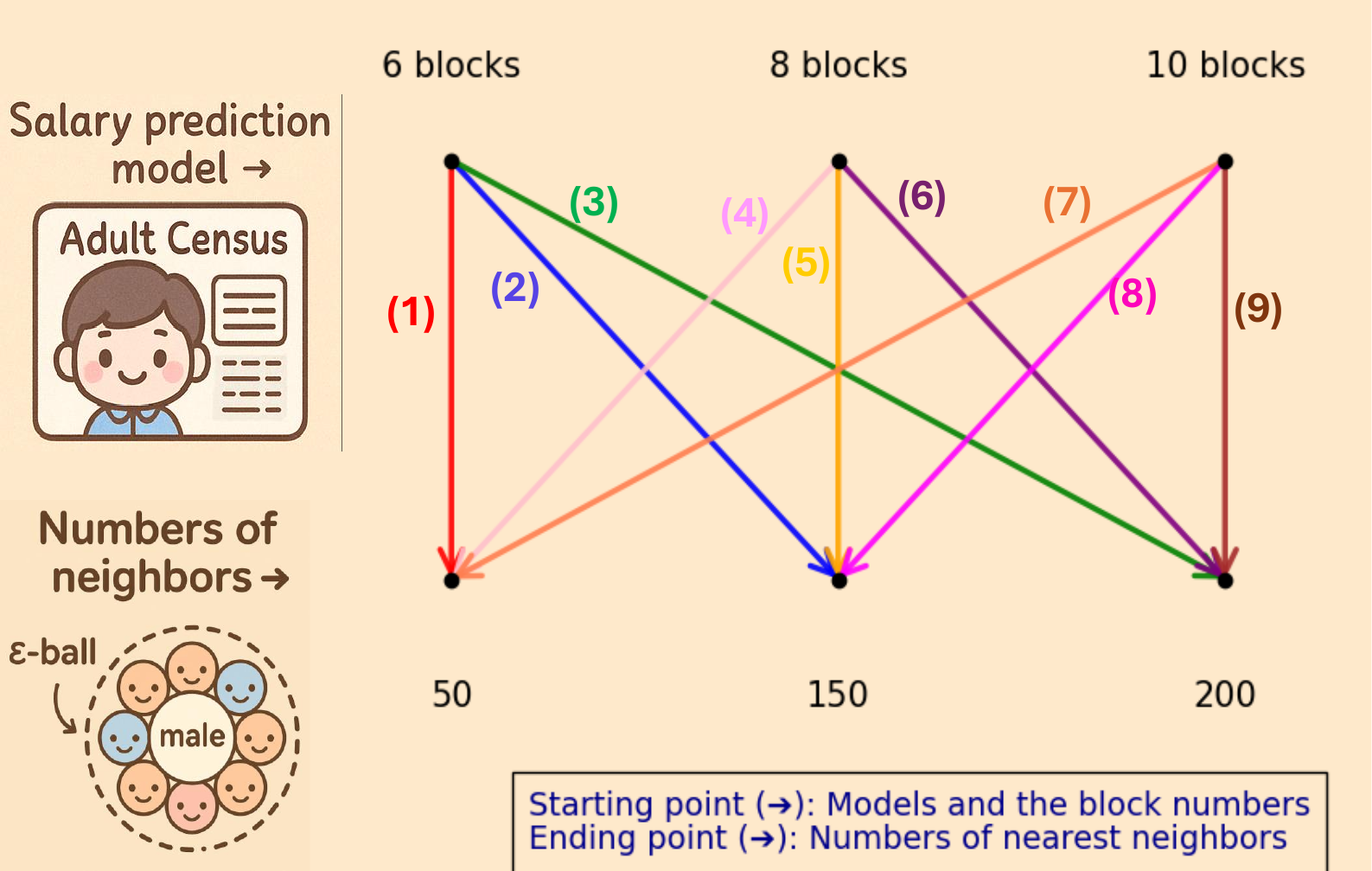}
    \caption{Ablation study on the influence of transformer block depth and neighborhood size on fairness, experiments (1) through (9) illustrate the different combinations of transformer block counts and neighborhood sizes evaluated in this section to assess their impact on fairness.
}
    \label{fig:abalation_block_neighbor}
\end{figure*}

In this section, we examine how fairness scores are influenced by varying combinations of transformer block counts and the number of nearest neighbors. The experimental configuration for this analysis is illustrated in \Cref{fig:abalation_block_neighbor}. As previously discussed in Section Experimental Results, we presented fairness scores for selected models with different transformer depths. Here, our objective is to provide a more comprehensive view, highlighting how fairness metrics evolve progressively as we adjust the number of transformer blocks and nearest neighbors. This analysis aims to uncover systematic trends and sensitivities within the model architecture that impact fairness.

For demonstration purposes, this analysis is limited to the salary prediction model within the context of gender bias mitigation verification, as our objective is to illustrate the influence of neighborhood size and the number of transformer blocks on changes in fairness scores, rather than to perform an exhaustive evaluation across all models. All of the hyperparameters used can be found in \Cref{sec:hyperparaneter_tuning}.

\paragraph{Fairness score for 6 blocks of transformer for salary prediction model with varying numbers of nearest neighbors.}

\begin{figure*}
    \centering
    \begin{minipage}[t]{0.31\textwidth}
        \centering
        \includegraphics[width=\linewidth]{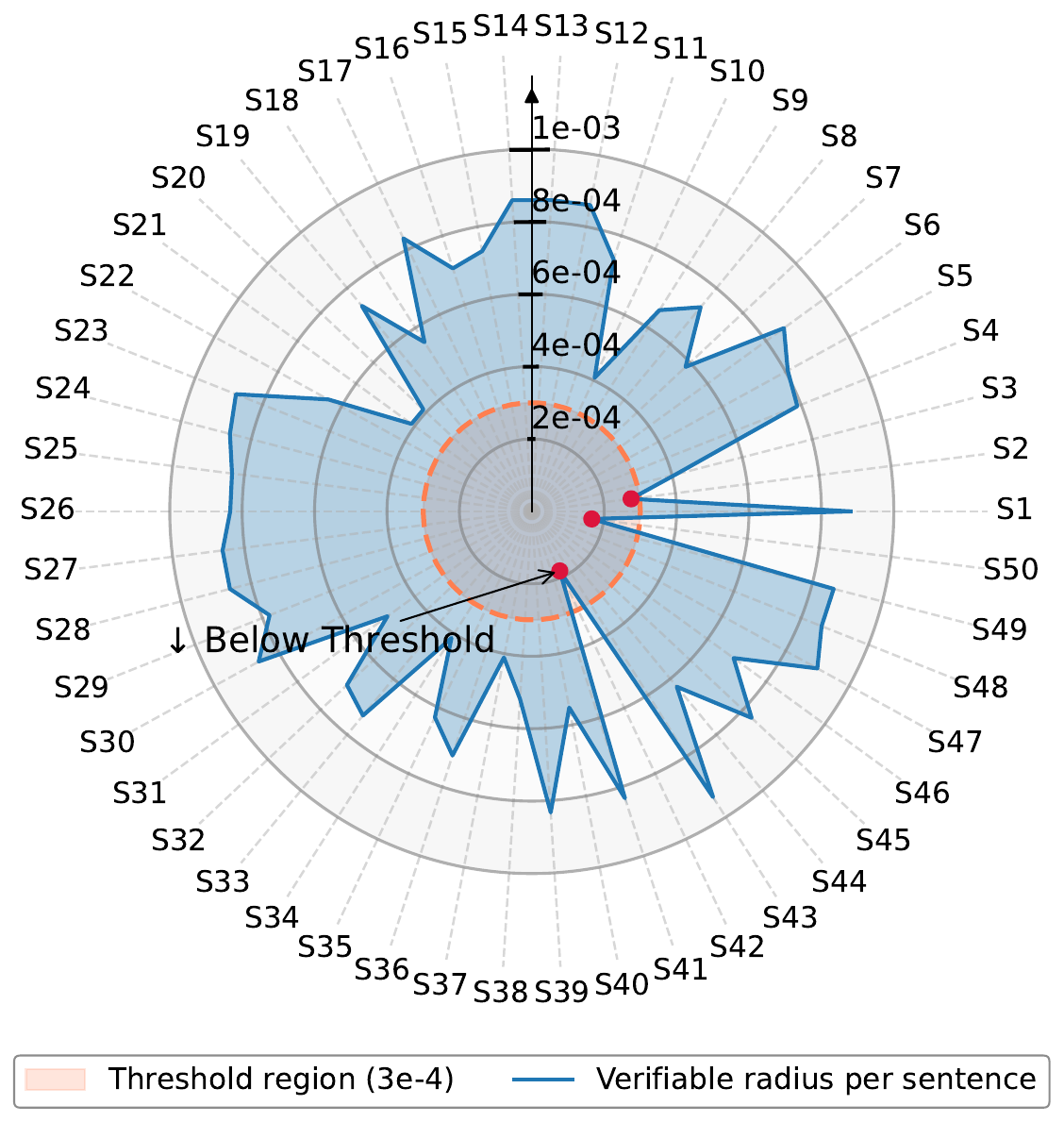}
        \caption{Fairness score for 6-block salary prediction model, 50 nearest neighbors (experiment (1) in \Cref{fig:abalation_block_neighbor}).}
        \label{fig:6b_50}
    \end{minipage}%
    \hfill
    \begin{minipage}[t]{0.31\textwidth}
        \centering
        \includegraphics[width=\linewidth]{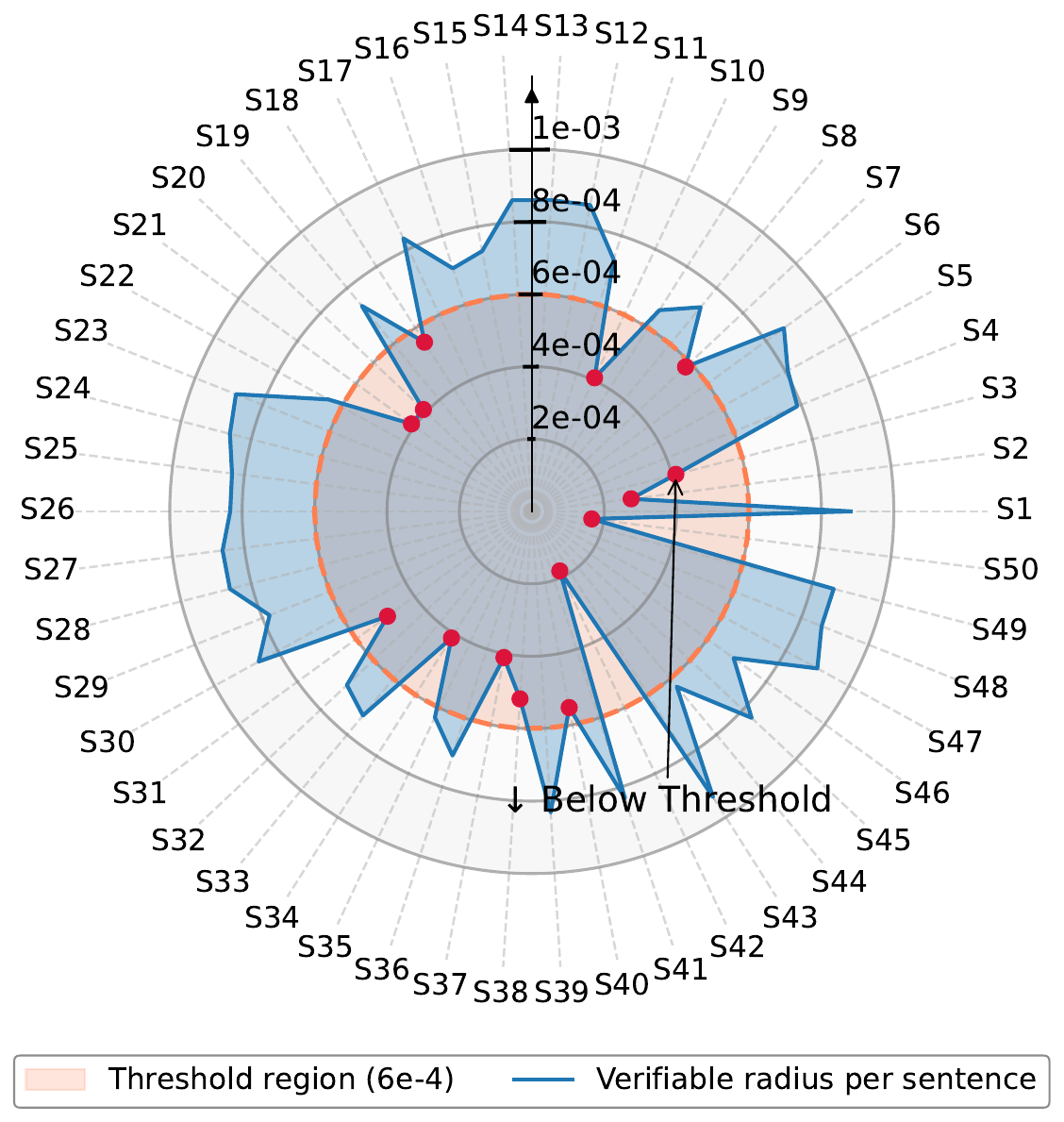}
        \caption{Fairness score for 6-block model, 150 nearest neighbors (experiment (2) in \Cref{fig:abalation_block_neighbor}).}
        \label{fig:6b_150}
    \end{minipage}%
    \hfill
    \begin{minipage}[t]{0.31\textwidth}
        \centering
        \includegraphics[width=\linewidth]{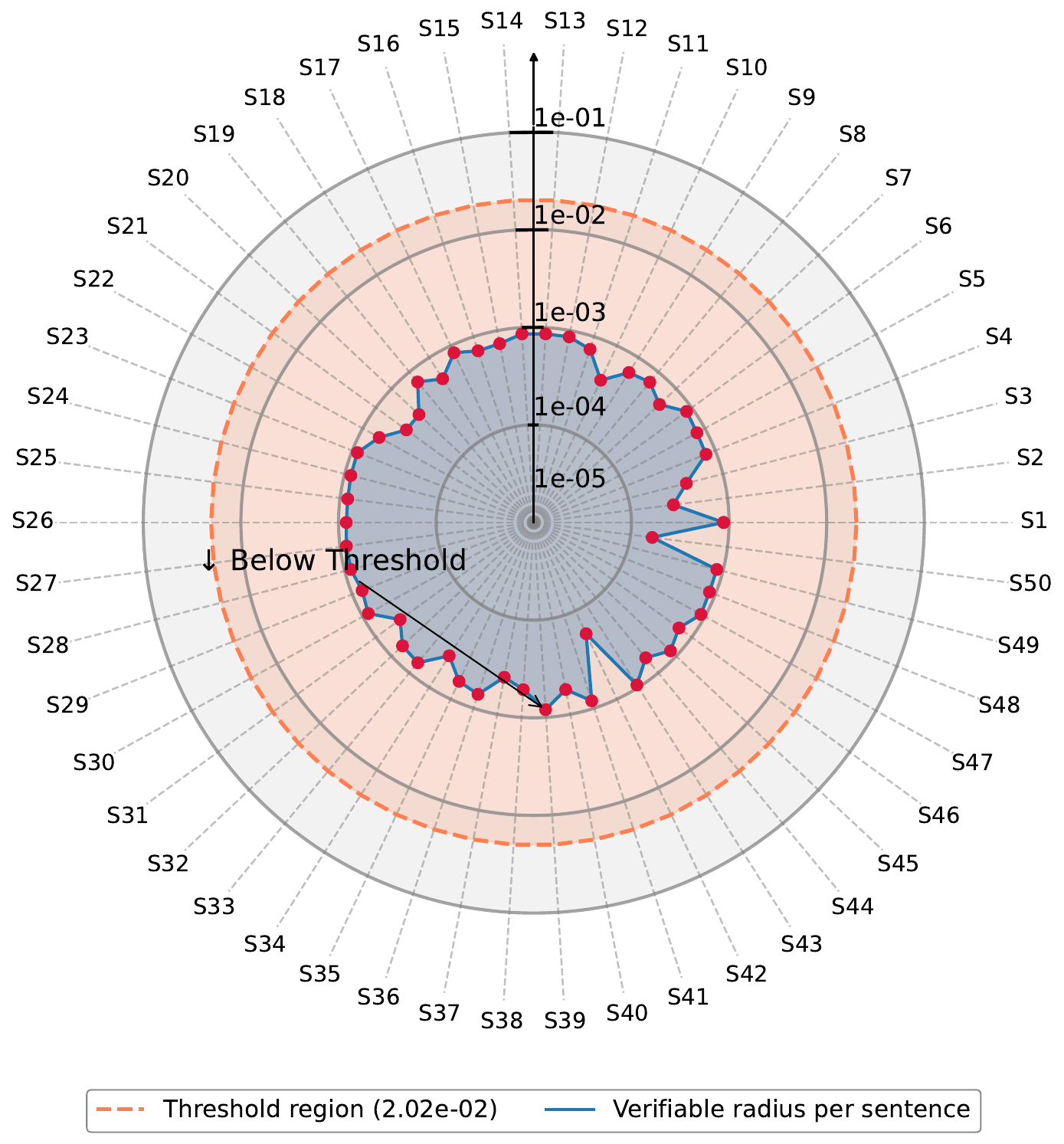}
        \caption{Fairness score for 6-block model, 200 nearest neighbors (experiment (3) in \Cref{fig:abalation_block_neighbor}).}
        \label{fig:6b_200}
    \end{minipage}

    \vspace{0.5cm}

    \begin{minipage}[t]{0.31\textwidth}
        \centering
        \includegraphics[width=\linewidth]{figures/verifiability_radar_8b_50.pdf}
        \caption{Verifiability radar plot with gender-distance reference circle for 8-block with 50 nearest neighbors (experiment (4) in \Cref{fig:abalation_block_neighbor}, also in Section Experimental Results).}
        \label{fig:verifiability_radar_8b_50}
    \end{minipage}%
    \hfill
    \begin{minipage}[t]{0.31\textwidth}
        \centering
        \includegraphics[width=\linewidth]{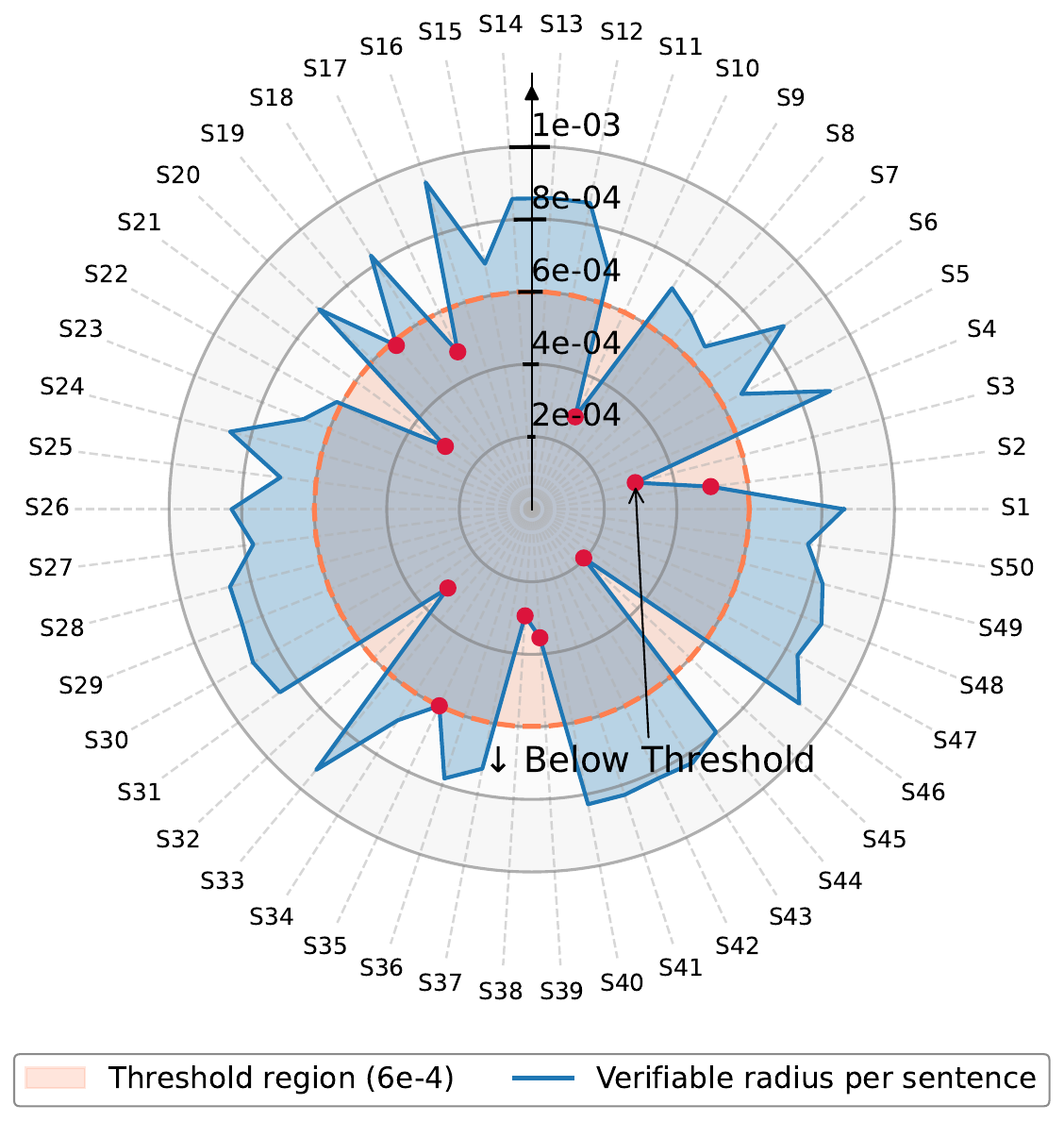}
        \caption{Verifiability radar plot with gender-distance reference circle for 8-block with 150 nearest neighbors (experiment (5) in \Cref{fig:abalation_block_neighbor}).}
        \label{fig:verifiability_radar_8b_150}
    \end{minipage}%
    \hfill
    \begin{minipage}[t]{0.31\textwidth}
        \centering
        \includegraphics[width=\linewidth]{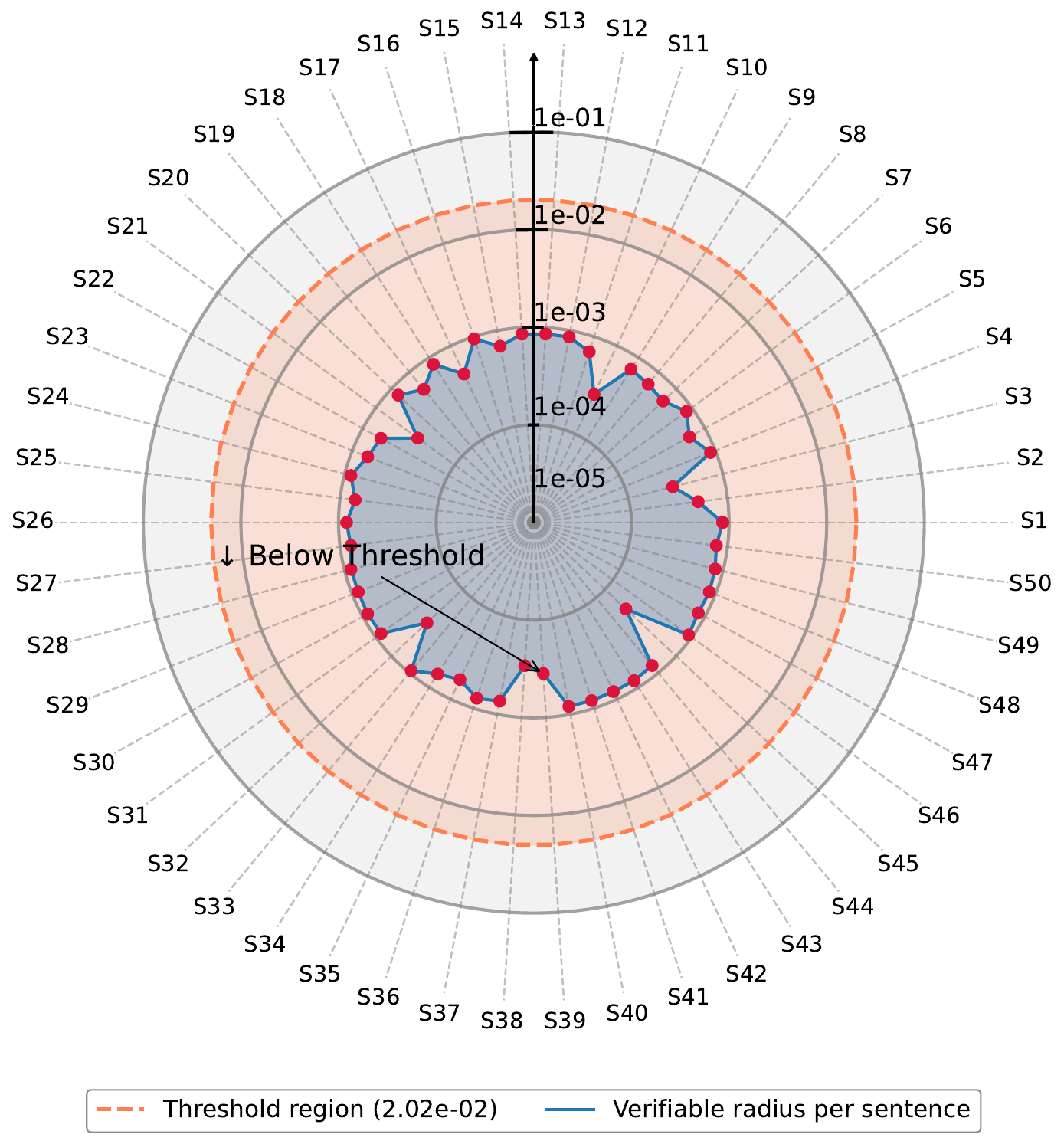}
        \caption{Verifiability radar plot with gender-distance reference circle for 8-block with 200 nearest neighbors (experiment (6) in \Cref{fig:abalation_block_neighbor}).}
        \label{fig:verifiability_radar_8b_200}
    \end{minipage}

    \vspace{0.5cm}

    \begin{minipage}[t]{0.31\textwidth}
        \centering
        \includegraphics[width=\linewidth]{figures/verifiability_radar_10b_50.pdf}
        \caption{Verifiability radar plot with gender-distance reference circle for 10-block with 50 nearest neighbors (experiment (7) in \Cref{fig:abalation_block_neighbor}, also in Section Experimental Results.}
        \label{fig:verifiability_radar_10b_50}
    \end{minipage}%
    \hfill
    \begin{minipage}[t]{0.31\textwidth}
        \centering
        \includegraphics[width=\linewidth]{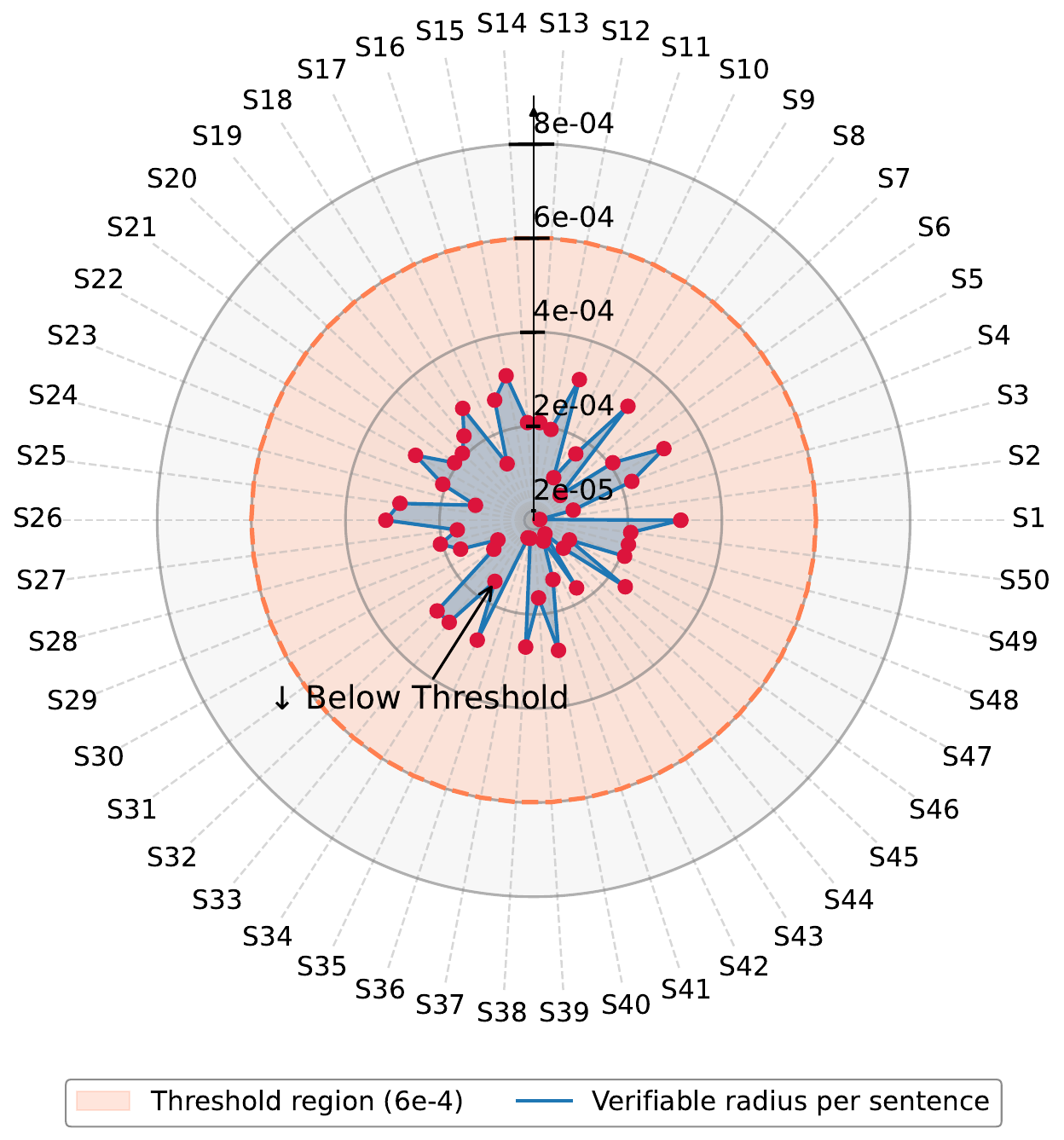}
        \caption{Verifiability radar plot with gender-distance reference circle for 10-block with 150 nearest neighbors (experiment (8) in \Cref{fig:abalation_block_neighbor}).}
        \label{fig:verifiability_radar_10b_150}
    \end{minipage}%
    \hfill
    \begin{minipage}[t]{0.31\textwidth}
        \centering
        \includegraphics[width=\linewidth]{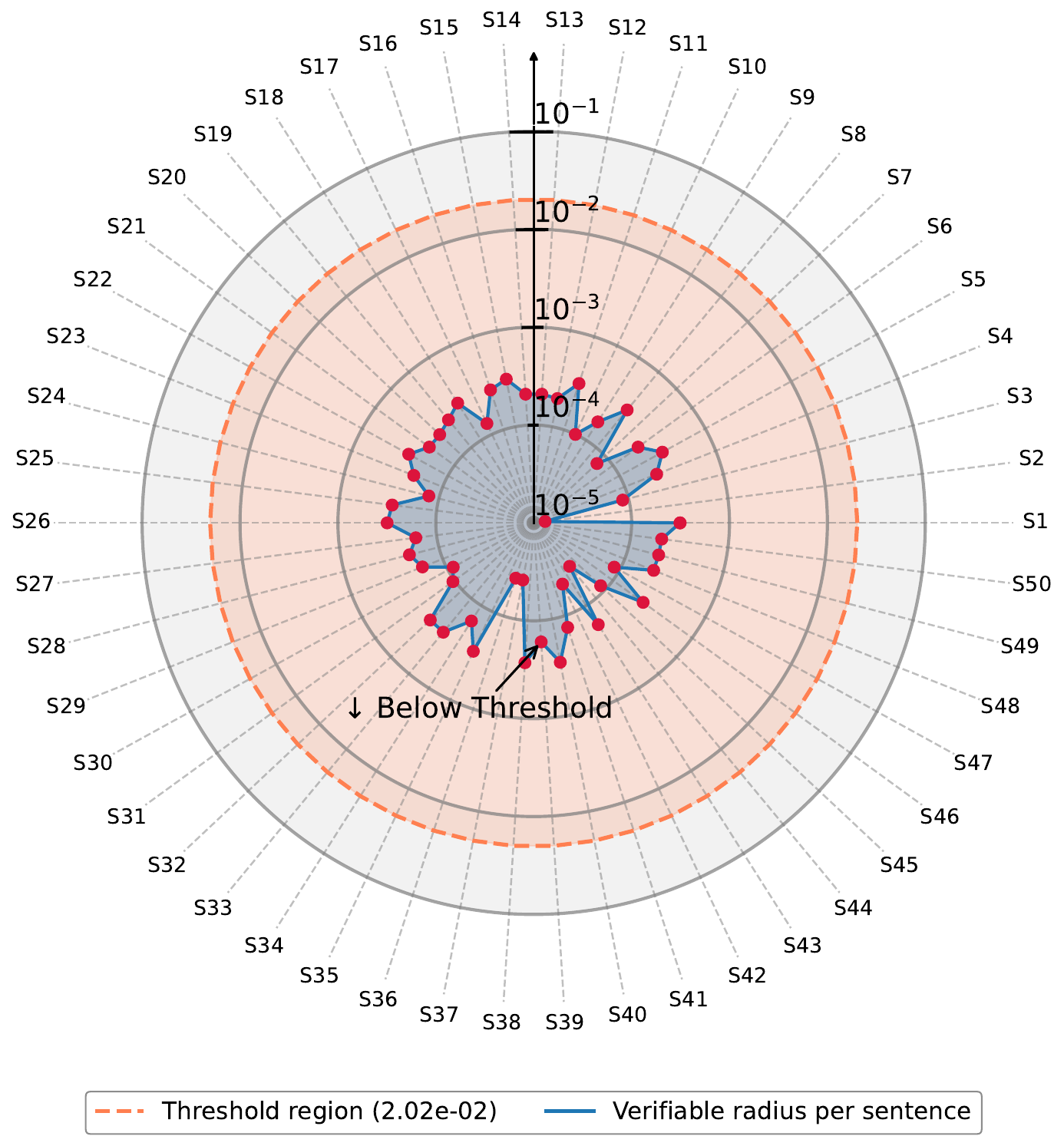}
        \caption{Verifiability radar plot with gender-distance reference circle for 10-block with 200 nearest neighbors (experiment (9) in \Cref{fig:abalation_block_neighbor}).}
        \label{fig:verifiability_radar_10b_200}
    \end{minipage}
\end{figure*}

We illustrate the progressive changes in fairness scores for the 6-block salary prediction model, evaluated across different settings of nearest neighbors (50, 150, and 200). These results highlight the impact of neighborhood size on fairness evaluation and are visualized in \Cref{fig:6b_50,fig:6b_150,fig:6b_200} (The selected numbers of neighbors are intended to illustrate the progressive changes that occur as neighborhood size increases. This specific choice (50, 150, 200) is justified by the observation that increasing the number of neighbors from 50 to 100 does not significantly increase the maximum distance between neighbors, nor does it result in noticeable changes in verifiability).

\paragraph{Fairness score for 8 blocks of transformer for salary prediction model with varying numbers of nearest neighbors.}

We illustrate the progressive changes in fairness scores for the 8-block salary prediction model, evaluated across different settings of nearest neighbors (50, 150, and 200). These results highlight the impact of neighborhood size on fairness evaluation and are visualized in \Cref{fig:verifiability_radar_8b_50,fig:verifiability_radar_8b_150,fig:verifiability_radar_8b_200}.

\paragraph{Fairness score for 10 blocks of transformer for salary prediction model with varying numbers of nearest neighbors.}

We illustrate the progressive changes in fairness scores for the 10-block salary prediction model, evaluated across different settings of nearest neighbors (50, 150, and 200). These results highlight the impact of neighborhood size on fairness evaluation and are visualized in \Cref{fig:verifiability_radar_10b_50,fig:verifiability_radar_10b_150,fig:verifiability_radar_10b_200}.

\subsubsection{Distances of words before and  after pre-training}
\label{sec:distance_complete}

This section presents the semantic relationships between words before and after the embedding model pre-training. The word pairs included are randomly selected to provide a representative illustration of the model's behavior.

The purpose of this section is to demonstrates the effectiveness of the pre-training process and, more importantly, ensures that the zonotope-based verification approach does not indiscriminately verify arbitrary words, but rather focuses on gendered terms within a sufficiently compact embedding space, This is achieved by ensuring that the verifiable noise radius is sufficiently large to include as many gendered words as possible, while small enough to exclude unrelated terms. This approach underscores the importance of careful hyperparameter tuning to balance coverage and precision in the verification process.

It should also be noted that while the proximity of gendered or toxicity-related words in the embedding space is largely influenced by the pulling factor ($\alpha$ in \eqref{eq:enhanced_contrastive_loss}), there appears to be a diminishing return in the Keras embedding layer’s ability to further cluster words from the same semantic class beyond a certain distance (2e-04), while also further separation of semantic classes yields minimal improvements.

\paragraph{Gendered words relationship before and after contrastive loss pre-training.}

The $\ell_2$ and $\ell_\infty$ distances between gendered word pairs, as well as between gendered and unrelated word pairs, before and after embedding pre-training are presented in \Cref{tab:distances_related_unrelated}. As shown, gendered word pairs become significantly closer in the embedding space after pre-training, while unrelated pairs are pushed farther apart, demonstrating the intended effect of the contrastive pre-training approach.

It is important to note that not all related word pairs exhibited reduced distances after pre-training, nor did all unrelated pairs show increased separation. This is not a limitation of the pre-training process itself, but rather a consequence of computational constraints, which prevented exhaustive pairing of all possible related and unrelated words; doing so would have significantly expanded the data constructed for pre-training and potentially degraded the efficiency and performance of the embedding model. Readers should note that the objective of this experiment (as well as for toxicity detection) is not to train the embedding model on an exhaustive set of related and unrelated word pairs, but rather to illustrate the effectiveness of the pre-training-based transfer learning approach in shaping semantically meaningful embeddings.

The 50 nearest neighbors of the gendered indicator words after pre-training in the Adult Census dataset are presented in \Cref{tab:gender_8b_pre}. A manual inspection was conducted to assess the relevance of these neighboring words to gender. The analysis confirmed that the majority of the neighbors are indeed gender-related. While a few terms—such as bull, stud, and colt—may not be explicitly gendered in all contexts, they can carry gendered connotations depending on usage. Overall, the presence of truly ungendered terms among the nearest neighbors is minimal and is determined by the authors to be negligible.

\begin{table*}
\centering

\setlength{\tabcolsep}{4pt}
\renewcommand{\arraystretch}{1.1}
\caption{Top 50 most $\ell_\infty$ -similar words to \token{female} and \token{male} using the Keras embedding layer after pre-training.}
\resizebox{\textwidth}{!}{
\begin{tabular}{lllllllll}
\toprule
\textbf{Model} & \multicolumn{8}{c}{\cellcolor{gray!10}\textbf{Salary Prediction \& Gender Bias Mitigation (8 \& 10 blocks)}} \\
\midrule
\textbf{} & \multicolumn{4}{c}{\textbf{\token{female}}} & \multicolumn{4}{c}{\textbf{\token{male}}} \\
\cmidrule(lr){2-5} \cmidrule(lr){6-9}
 & \textbf{Word} & \textbf{Distance ($\downarrow$)} & \textbf{Word} & \textbf{Distance ($\downarrow$)} & \textbf{Word} & \textbf{Distance ($\downarrow$)} & \textbf{Word} & \textbf{Distance ($\downarrow$)} \\
\midrule
\multirow{25}{*}{} 
 & \token{nun}         & $2.0\text{e}{-04}$ & \token{son}        & $2.0\text{e}{-04}$ & \token{actress}      & $2.0\text{e}{-04}$ & \token{woman}        & $2.0\text{e}{-04}$ \\
 & \token{deer}        & $2.0\text{e}{-04}$ & \token{busch}      & $2.0\text{e}{-04}$ & \token{busch}        & $2.0\text{e}{-04}$ & \token{her}          & $2.0\text{e}{-04}$ \\
 & \token{gal}         & $2.0\text{e}{-04}$ & \token{bride}      & $2.0\text{e}{-04}$ & \token{hostess}      & $2.0\text{e}{-04}$ & \token{goddess}      & $2.0\text{e}{-04}$ \\
 & \token{bull}        & $2.0\text{e}{-04}$ & \token{dad}        & $2.0\text{e}{-04}$ & \token{gonzales}     & $2.0\text{e}{-04}$ & \token{dude}         & $2.0\text{e}{-04}$ \\
 & \token{brothers}    & $2.0\text{e}{-04}$ & \token{lads}       & $2.0\text{e}{-04}$ & \token{bro}          & $2.0\text{e}{-04}$ & \token{maid}         & $2.0\text{e}{-04}$ \\
 & \token{grandpa}     & $2.0\text{e}{-04}$ & \token{bro}        & $2.0\text{e}{-04}$ & \token{brother}      & $2.0\text{e}{-04}$ & \token{mama}         & $2.0\text{e}{-04}$ \\
 & \token{earl}        & $2.0\text{e}{-04}$ & \token{hostess}    & $2.0\text{e}{-04}$ & \token{chick}        & $2.0\text{e}{-04}$ & \token{godfather}    & $2.0\text{e}{-04}$ \\
 & \token{sister}      & $2.0\text{e}{-04}$ & \token{grandfather}& $2.0\text{e}{-04}$ & \token{womb}         & $2.0\text{e}{-04}$ & \token{feminist}     & $2.0\text{e}{-04}$ \\
 & \token{momma}       & $2.0\text{e}{-04}$ & \token{maid}       & $2.0\text{e}{-04}$ & \token{fiancee}      & $2.0\text{e}{-04}$ & \token{granny}       & $2.0\text{e}{-04}$ \\
 & \token{lion}        & $2.0\text{e}{-04}$ & \token{fiance}     & $2.0\text{e}{-04}$ & \token{earl}         & $2.0\text{e}{-04}$ & \textbf{\token{female}} & \textbf{$3.0\text{e}{-04}$} \\
 & \token{colts}       & $2.0\text{e}{-04}$ & \textbf{\token{male}} & \textbf{$3.0\text{e}{-04}$} & \token{guy}      & $3.0\text{e}{-04}$ & \token{businessman}  & $3.0\text{e}{-04}$ \\
 & \token{brother}     & $3.0\text{e}{-04}$ & \token{boy}        & $3.0\text{e}{-04}$ & \token{brothers}     & $3.0\text{e}{-04}$ & \token{colts}        & $3.0\text{e}{-04}$ \\
 & \token{widow}       & $3.0\text{e}{-04}$ & \token{womb}       & $3.0\text{e}{-04}$ & \token{lion}         & $3.0\text{e}{-04}$ & \token{chairman}     & $3.0\text{e}{-04}$ \\
 & \token{stallion}    & $3.0\text{e}{-04}$ & \token{her}        & $3.0\text{e}{-04}$ & \token{queen}        & $3.0\text{e}{-04}$ & \token{she}          & $3.0\text{e}{-04}$ \\
 & \token{fiancee}     & $3.0\text{e}{-04}$ & \token{godfather}  & $3.0\text{e}{-04}$ & \token{monk}         & $3.0\text{e}{-04}$ & \token{wife}         & $3.0\text{e}{-04}$ \\
 & \token{father}      & $3.0\text{e}{-04}$ & \token{mother}     & $3.0\text{e}{-04}$ & \token{stud}         & $3.0\text{e}{-04}$ & \token{males}        & $3.0\text{e}{-04}$ \\
 & \token{grandmother} & $3.0\text{e}{-04}$ & \token{madame}     & $3.0\text{e}{-04}$ & \token{granddaughter}& $3.0\text{e}{-04}$ & \token{diva}         & $3.0\text{e}{-04}$ \\
 & \token{statesman}   & $3.0\text{e}{-04}$ & \token{dude}       & $3.0\text{e}{-04}$ & \token{lads}         & $3.0\text{e}{-04}$ & \token{nuns}         & $3.0\text{e}{-04}$ \\
 & \token{nuns}        & $3.0\text{e}{-04}$ & \token{gentlemen}  & $3.0\text{e}{-04}$ & \token{deer}         & $3.0\text{e}{-04}$ & \token{man}          & $3.0\text{e}{-04}$ \\
 & \token{businessman} & $3.0\text{e}{-04}$ & \token{guy}        & $3.0\text{e}{-04}$ & \token{mother}       & $3.0\text{e}{-04}$ & \token{papa}         & $3.0\text{e}{-04}$ \\
 & \token{aunt}        & $3.0\text{e}{-04}$ & \token{him}        & $3.0\text{e}{-04}$ & \token{women}        & $3.0\text{e}{-04}$ & \token{kings}        & $3.0\text{e}{-04}$ \\
 & \token{maternal}    & $3.0\text{e}{-04}$ & \token{fathers}    & $3.0\text{e}{-04}$ & \token{statesman}    & $3.0\text{e}{-04}$ & \token{his}          & $3.0\text{e}{-04}$ \\
 & \token{himself}     & $3.0\text{e}{-04}$ & \token{queen}      & $3.0\text{e}{-04}$ & \token{himself}      & $3.0\text{e}{-04}$ & \token{spokesman}    & $3.0\text{e}{-04}$ \\
 & \token{mama}        & $3.0\text{e}{-04}$ & \token{men}        & $3.0\text{e}{-04}$ & \token{son}          & $3.0\text{e}{-04}$ & \token{sons}         & $3.0\text{e}{-04}$ \\
 & \token{wife}        & $3.0\text{e}{-04}$ &                        &                    &                        &                    &                        &                    \\

\bottomrule
\end{tabular}}

\label{tab:gender_8b_pre}
\end{table*}

\begin{table*}
\centering

\setlength{\tabcolsep}{5pt}
\renewcommand{\arraystretch}{1.15}
\caption{$\ell_2$  and $\ell_\infty$  distances between word pairs before and after embedding pre-training. Related gendered words become closer, while unrelated pairs grow further apart.}
\begin{tabular}{l c l c c c c}
\toprule
 &\textbf{Relation} & \multicolumn{5}{c}{\cellcolor{gray!10}\textbf{Related gendered word pairs}} \\
\midrule
\textbf{} & \textbf{Word pair} &  &\textbf{$\ell_2$  (Before)} & \textbf{$\ell_2$  (After) ($\downarrow$)} & \textbf{$\ell_\infty$  (Before)} & \textbf{$\ell_\infty$  (After) ($\downarrow$)} \\
\midrule
\token{man}     & $\leftrightarrow$ & \token{woman}       & $3.22\text{e}{-02}$ & $5.00\text{e}{-04}$ & $7.59\text{e}{-02}$ & $4.00\text{e}{-04}$ \\
\token{king}    & $\leftrightarrow$ & \token{queen}       & $1.99\text{e}{-02}$ & $7.00\text{e}{-04}$ & $4.94\text{e}{-02}$ & $4.00\text{e}{-04}$ \\
\token{boy}     & $\leftrightarrow$ & \token{girl}        & $2.89\text{e}{-02}$ & $9.00\text{e}{-04}$ & $6.36\text{e}{-02}$ & $5.00\text{e}{-04}$ \\
\token{female}  & $\leftrightarrow$ & \token{male}        & $1.37\text{e}{-02}$ & $4.00\text{e}{-04}$ & $3.27\text{e}{-02}$ & $3.00\text{e}{-04}$ \\
\token{father}  & $\leftrightarrow$ & \token{mother}      & $2.77\text{e}{-02}$ & $5.00\text{e}{-04}$ & $6.93\text{e}{-02}$ & $2.00\text{e}{-04}$ \\
\token{husband} & $\leftrightarrow$ & \token{wife}        & $1.98\text{e}{-02}$ & $8.00\text{e}{-04}$ & $4.92\text{e}{-02}$ & $6.00\text{e}{-04}$ \\
\token{brother} & $\leftrightarrow$ & \token{sister}      & $3.05\text{e}{-02}$ & $6.00\text{e}{-04}$ & $7.54\text{e}{-02}$ & $4.00\text{e}{-04}$ \\
\token{son}     & $\leftrightarrow$ & \token{daughter}    & $2.59\text{e}{-02}$ & $5.00\text{e}{-04}$ & $5.66\text{e}{-02}$ & $3.00\text{e}{-04}$ \\
\token{uncle}   & $\leftrightarrow$ & \token{aunt}        & $2.91\text{e}{-02}$ & $8.00\text{e}{-04}$ & $7.21\text{e}{-02}$ & $5.00\text{e}{-04}$ \\
\token{nephew}  & $\leftrightarrow$ & \token{niece}       & $2.78\text{e}{-02}$ & $7.00\text{e}{-04}$ & $6.70\text{e}{-02}$ & $4.00\text{e}{-04}$ \\
\token{prince}  & $\leftrightarrow$ & \token{princess}    & $2.50\text{e}{-02}$ & $7.00\text{e}{-04}$ & $5.92\text{e}{-02}$ & $5.00\text{e}{-04}$ \\
\token{monk}    & $\leftrightarrow$ & \token{nun}         & $3.49\text{e}{-02}$ & $8.00\text{e}{-04}$ & $8.63\text{e}{-02}$ & $7.00\text{e}{-04}$ \\
\token{wizard}  & $\leftrightarrow$ & \token{witch}       & $3.33\text{e}{-02}$ & $7.55\text{e}{-02}$ & $8.20\text{e}{-02}$ & $5.38\text{e}{-02}$ \\
\token{female}  & $\leftrightarrow$ & \token{grandpa}     & $3.39\text{e}{-02}$ & $4.00\text{e}{-04}$ & $8.25\text{e}{-02}$ & $2.00\text{e}{-04}$ \\
\bottomrule

\end{tabular}

\begin{tabular}{l c l c c c c}
\toprule
 &\textbf{Relation} & \multicolumn{5}{c}{\cellcolor{gray!10}\textbf{Unrelated word pairs}} \\
\midrule
\textbf{} & \textbf{Word pair} & &\textbf{$\ell_2$  (Before)} & \textbf{$\ell_2$  (After) ($\uparrow$)} & \textbf{$\ell_\infty$  (Before)} & \textbf{$\ell_\infty$  (After) ($\uparrow$)} \\
\midrule
\token{woman}     & $\leftrightarrow$ & \token{ocean}       & $1.81\text{e}{-02}$ & $2.381\text{e}{-01}$ & $4.37\text{e}{-02}$ & $1.360\text{e}{-01}$ \\
\token{boy}       & $\leftrightarrow$ & \token{guitar}      & $2.10\text{e}{-02}$ & $6.659\text{e}{-01}$ & $4.68\text{e}{-02}$ & $4.209\text{e}{-01}$ \\
\token{girl}      & $\leftrightarrow$ & \token{airplane}    & $2.33\text{e}{-02}$ & $1.557\text{e}{-01}$ & $5.60\text{e}{-02}$ & $1.033\text{e}{-01}$ \\
\token{father}    & $\leftrightarrow$ & \token{bottle}      & $3.11\text{e}{-02}$ & $1.0779\text{e}{+00}$ & $7.97\text{e}{-02}$ & $4.445\text{e}{-01}$ \\
\token{mother}    & $\leftrightarrow$ & \token{keyboard}    & $2.61\text{e}{-02}$ & $5.032\text{e}{-01}$ & $6.79\text{e}{-02}$ & $2.245\text{e}{-01}$ \\
\token{husband}   & $\leftrightarrow$ & \token{river}       & $3.14\text{e}{-02}$ & $4.214\text{e}{-01}$ & $6.76\text{e}{-02}$ & $2.457\text{e}{-01}$ \\
\token{wife}      & $\leftrightarrow$ & \token{planet}      & $2.63\text{e}{-02}$ & $2.126\text{e}{-01}$ & $5.36\text{e}{-02}$ & $1.400\text{e}{-01}$ \\
\token{brother}   & $\leftrightarrow$ & \token{window}      & $2.30\text{e}{-02}$ & $8.35\text{e}{-02}$ & $5.72\text{e}{-02}$ & $6.52\text{e}{-02}$ \\
\token{sister}    & $\leftrightarrow$ & \token{chair}       & $3.07\text{e}{-02}$ & $8.44\text{e}{-02}$ & $7.77\text{e}{-02}$ & $5.12\text{e}{-02}$ \\
\token{son}       & $\leftrightarrow$ & \token{clock}       & $3.26\text{e}{-02}$ & $1.1346\text{e}{+00}$ & $7.90\text{e}{-02}$ & $5.048\text{e}{-01}$ \\
\token{daughter}  & $\leftrightarrow$ & \token{road}        & $1.43\text{e}{-02}$ & $4.082\text{e}{-01}$ & $3.01\text{e}{-02}$ & $2.651\text{e}{-01}$ \\
\token{uncle}     & $\leftrightarrow$ & \token{painting}    & $2.46\text{e}{-02}$ & $7.58\text{e}{-02}$ & $5.71\text{e}{-02}$ & $4.90\text{e}{-02}$ \\
\token{aunt}      & $\leftrightarrow$ & \token{garden}      & $1.34\text{e}{-02}$ & $3.049\text{e}{-01}$ & $2.94\text{e}{-02}$ & $1.480\text{e}{-01}$ \\
\bottomrule
\end{tabular}

\label{tab:distances_related_unrelated}
\end{table*}

\paragraph{Toxic words relationship before and after contrastive loss pre-training.}

The $\ell_2$ and $\ell_\infty$ distances between toxic word pairs, as well as between toxic and unrelated word pairs, before and after embedding pre-training are presented in \Cref{tab:toxicity_related_unrelated}. As shown, toxicity-related word pairs become significantly closer in the embedding space after pre-training, while unrelated pairs are pushed farther apart, demonstrating the intended effect of the contrastive pre-training approach.

The 50 nearest neighbors of some of one of the toxicity-related words (\token{bitch}) are presented in \Cref{tab:prof_50_nearest} as an example. A manual inspection was conducted to assess the relevance of these neighboring words to this specific toxicity-related word. The analysis confirmed that the majority of the neighbors are indeed toxicity-related.

\begin{table*}
\centering
\small
\setlength{\tabcolsep}{6pt}
\renewcommand{\arraystretch}{1.2}
\caption{Top 50 most $\ell_\infty$ -similar words to \token{bitch} using the pre-trained Keras embedding layer.}
\begin{tabular}{lll ll ll}
\toprule
&\textbf{Relation} &  \multicolumn{5}{c}{\cellcolor{gray!10}\textbf{Similar words  to \token{bitch} ($\ell_\infty$  distance)}} \\
\midrule
\textbf{} & \textit{[PAD]}       & $1.645\text{e}{-03}$ & \token{\#\#tter}   & $2.569\text{e}{-03}$ & \token{cock}     & $2.631\text{e}{-03}$ \\
\textbf{} & \token{\#\#ter}   & $1.897\text{e}{-03}$ & \token{screwing}   & $2.686\text{e}{-03}$ & \token{shy}      & $2.710\text{e}{-03}$ \\
\textbf{} & \token{\#\#kker}  & $2.267\text{e}{-03}$ & \token{\#\#s}      & $2.804\text{e}{-03}$ & \token{\#\#holz} & $2.854\text{e}{-03}$ \\
\textbf{} & \token{\#\#r}     & $2.320\text{e}{-03}$ & \token{pack}       & $2.856\text{e}{-03}$ & \token{\#\#1}    & $3.011\text{e}{-03}$ \\
\textbf{} & \token{\#\#z}     & $2.409\text{e}{-03}$ & \token{wan}        & $3.100\text{e}{-03}$ & \token{\#\#ie}   & $3.137\text{e}{-03}$ \\
\textbf{} & \token{bi}        & $2.445\text{e}{-03}$ & \token{\#\#t}      & $3.145\text{e}{-03}$ & \token{;}        & $3.154\text{e}{-03}$ \\
\textbf{} & \token{\#\#cks}   & $2.533\text{e}{-03}$ & \token{*}          & $3.168\text{e}{-03}$ & \token{\#\#tt}   & $3.179\text{e}{-03}$ \\
\textbf{} & \token{\#\#us}    & $3.188\text{e}{-03}$ & \token{\#\#3}      & $3.261\text{e}{-03}$ & \token{\#\#ra}   & $3.289\text{e}{-03}$ \\
\textbf{} & \token{+}         & $3.340\text{e}{-03}$ & \token{fu}         & $3.386\text{e}{-03}$ & \token{\#\#te}   & $3.416\text{e}{-03}$ \\
\textbf{} & \token{(}         & $3.431\text{e}{-03}$ & \token{\#\#i}      & $3.469\text{e}{-03}$ & \token{\#\#ot}   & $3.471\text{e}{-03}$ \\
\textbf{} & \token{\#\#or}    & $3.496\text{e}{-03}$ & \token{\#\#er}     & $3.502\text{e}{-03}$ & \token{mother}   & $3.514\text{e}{-03}$ \\
\textbf{} & \token{\#\#ts}    & $3.632\text{e}{-03}$ & \token{knob}       & $3.662\text{e}{-03}$ & \token{ti}       & $3.757\text{e}{-03}$ \\
\textbf{} & \token{\#\#dge}   & $3.806\text{e}{-03}$ & \token{\#\#rac}    & $3.837\text{e}{-03}$ & \token{ass}      & $3.874\text{e}{-03}$ \\
\textbf{} & \token{moth}      & $3.899\text{e}{-03}$ & \token{pick}       & $3.905\text{e}{-03}$ & \token{fuck}     & $3.937\text{e}{-03}$ \\
\textbf{} & \token{sucker}    & $3.976\text{e}{-03}$ & \token{\#\#ey}     & $3.987\text{e}{-03}$ & \token{\#\#ore}  & $4.006\text{e}{-03}$ \\
\textbf{} & \token{ho}        & $4.020\text{e}{-03}$ & \token{ku}         & $4.116\text{e}{-03}$ & \token{-}        & $4.118\text{e}{-03}$ \\
\textbf{} & \token{!}         & $4.165\text{e}{-03}$ & \token{\#\#it}     & $4.184\text{e}{-03}$ 
\\
\bottomrule
\end{tabular}

\label{tab:prof_50_nearest}
\end{table*}

\paragraph{Semantic relationships within general vocabulary terms.}

A selection of randomly chosen synonym and antonym pairs, along with their semantic embedding distances before and after embedding pre-training, is presented in \Cref{tab:distance_metrics_gen_sen}. As observed, synonym pairs exhibit reduced distances while antonym pairs become more distant in the embedding space after pre-training, demonstrating the intended effect of the contrastive learning approach.

\begin{table*}
\centering
\small{

\renewcommand{\arraystretch}{1.15}
\caption{$\ell_2$ and $\ell_\infty$ distances between synonym and antonym word pairs before and after embedding pre-training. Synonym pairs become closer, while antonym pairs grow further apart.}
\begin{tabularx}{\textwidth}{l c l *{4}{X}}
\toprule
&\textbf{Relation} & & \multicolumn{4}{c}{\cellcolor{gray!10}\textbf{Synonym word pairs}} \\
\midrule
 & \textbf{Word pair}& &$\ell_2$  (Before) & $\ell_2$  (After) ($\downarrow$) & $\ell_\infty$  (Before) & $\ell_\infty$  (After) ($\downarrow$) \\
\midrule
\token{big}         & $\leftrightarrow$ & \token{large}       & $2.22\text{e}{-02}$ & $5.50\text{e}{-03}$ & $1.45\text{e}{-02}$ & $2.70\text{e}{-03}$ \\
\token{smart}       & $\leftrightarrow$ & \token{intelligent} & $3.30\text{e}{-02}$ & $1.50\text{e}{-02}$ & $1.92\text{e}{-02}$ & $8.60\text{e}{-03}$ \\
\token{sad}         & $\leftrightarrow$ & \token{unhappy}     & $3.47\text{e}{-02}$ & $1.68\text{e}{-02}$ & $1.79\text{e}{-02}$ & $9.60\text{e}{-03}$ \\
\token{start}       & $\leftrightarrow$ & \token{begin}       & $2.78\text{e}{-02}$ & $7.60\text{e}{-03}$ & $1.81\text{e}{-02}$ & $4.40\text{e}{-03}$ \\
\bottomrule
\end{tabularx}

\vspace{0.1em}

\begin{tabularx}{\textwidth}{l c l *{4}{X}}
\toprule
&\textbf{Relation} & & \multicolumn{4}{c}{\cellcolor{gray!10}\textbf{Antonym word pairs}} \\
\midrule
&\textbf{Word pair} & & $\ell_2$  (Before) & $\ell_2$  (After) ($\uparrow$) & $\ell_\infty$  (Before) & $\ell_\infty$  (After) ($\uparrow$) \\
\midrule
\token{fast}        & $\leftrightarrow$ & \token{slow}        & $3.33\text{e}{-02}$ & $3.236\text{e}{-01}$ & $1.86\text{e}{-02}$ & $1.522\text{e}{-01}$ \\
\token{light}       & $\leftrightarrow$ & \token{dark}        & $2.37\text{e}{-02}$ & $2.741\text{e}{-01}$ & $1.42\text{e}{-02}$ & $1.273\text{e}{-01}$ \\
\token{love}        & $\leftrightarrow$ & \token{hate}        & $2.16\text{e}{-02}$ & $3.820\text{e}{-01}$ & $1.49\text{e}{-02}$ & $2.134\text{e}{-01}$ \\
\token{up}          & $\leftrightarrow$ & \token{down}        & $2.96\text{e}{-02}$ & $8.497\text{e}{-01}$ & $2.10\text{e}{-02}$ & $4.821\text{e}{-01}$ \\
\bottomrule
\end{tabularx}
}

\label{tab:distance_metrics_gen_sen}
\end{table*}

\begin{table*}
\centering
\small{
\caption{$\ell_2$ and $\ell_\infty$ distances between word pairs before and after embedding pre-training. Related toxic words become closer, while unrelated pairs grow further apart.}
\setlength{\tabcolsep}{5pt}
\renewcommand{\arraystretch}{1.15}
\resizebox{\textwidth}{!}{%
\begin{tabular}{>{\raggedleft\arraybackslash}p{0.23\textwidth} @{\hskip 2pt} c @{\hskip 2pt} >{\raggedright\arraybackslash}p{0.23\textwidth} cccc}
\toprule
 &\textbf{Relation} & & \multicolumn{4}{c}{\cellcolor{gray!10}\textbf{Related toxic word pairs}} \\
\midrule
 &  \textbf{Word pair}& & $\ell_2$ (Before) & $\ell_2$ (After) ($\downarrow$) & $\ell_\infty$ (Before) & $\ell_\infty$ (After) ($\downarrow$) \\

\midrule
\token{Mother Fuker} & $\leftrightarrow$ & \token{mother-fucker} & $7.43\text{e}{-02}$ & $2.70\text{e}{-03}$ & $1.806\text{e}{-01}$ & $1.60\text{e}{-03}$ \\
\token{blowjob} & $\leftrightarrow$ & \token{b17ch} & $5.18\text{e}{-02}$ & $1.90\text{e}{-03}$ & $1.328\text{e}{-01}$ & $1.20\text{e}{-03}$ \\
\token{gayboy} & $\leftrightarrow$ & \token{gays} & $2.23\text{e}{-02}$ & $2.70\text{e}{-03}$ & $5.34\text{e}{-02}$ & $1.60\text{e}{-03}$ \\
\token{dick*} & $\leftrightarrow$ & \token{dirsa} & $3.51\text{e}{-02}$ & $5.60\text{e}{-03}$ & $8.74\text{e}{-02}$ & $4.00\text{e}{-03}$ \\
\token{Phukker} & $\leftrightarrow$ & \token{peceenusss} & $5.43\text{e}{-02}$ & $4.20\text{e}{-03}$ & $1.168\text{e}{-01}$ & $2.30\text{e}{-03}$ \\
\token{puta} & $\leftrightarrow$ & \token{pusse} & $2.42\text{e}{-02}$ & $2.60\text{e}{-03}$ & $5.64\text{e}{-02}$ & $1.70\text{e}{-03}$ \\
\token{arse*} & $\leftrightarrow$ & \token{azzhole} & $2.74\text{e}{-02}$ & $3.10\text{e}{-03}$ & $6.95\text{e}{-02}$ & $2.30\text{e}{-03}$ \\
\token{chuj} & $\leftrightarrow$ & \token{cum} & $5.74\text{e}{-02}$ & $2.70\text{e}{-03}$ & $1.318\text{e}{-01}$ & $2.50\text{e}{-03}$ \\
\token{c0cks} & $\leftrightarrow$ & \token{cunt*} & $5.91\text{e}{-02}$ & $2.50\text{e}{-03}$ & $1.352\text{e}{-01}$ & $1.60\text{e}{-03}$ \\
\token{shits} & $\leftrightarrow$ & \token{sh1t} & $6.75\text{e}{-02}$ & $2.50\text{e}{-03}$ & $1.478\text{e}{-01}$ & $1.50\text{e}{-03}$ \\
\token{skanks} & $\leftrightarrow$ & \token{slutty} & $4.73\text{e}{-02}$ & $3.50\text{e}{-03}$ & $1.069\text{e}{-01}$ & $1.80\text{e}{-03}$ \\
\token{fuckin} & $\leftrightarrow$ & \token{faget} & $3.44\text{e}{-02}$ & $3.20\text{e}{-03}$ & $8.28\text{e}{-02}$ & $2.20\text{e}{-03}$ \\
\token{cawks} & $\leftrightarrow$ & \token{cock} & $8.18\text{e}{-02}$ & $3.10\text{e}{-03}$ & $2.174\text{e}{-01}$ & $1.90\text{e}{-03}$ \\
\token{niiger} & $\leftrightarrow$ & \token{nigger} & $3.25\text{e}{-02}$ & $2.70\text{e}{-03}$ & $8.01\text{e}{-02}$ & $1.80\text{e}{-03}$ \\
\token{asshole} & $\leftrightarrow$ & \token{assface} & $5.45\text{e}{-02}$ & $3.00\text{e}{-03}$ & $1.339\text{e}{-01}$ & $2.70\text{e}{-03}$ \\
\token{sh1t} & $\leftrightarrow$ & \token{screw} & $7.62\text{e}{-02}$ & $3.60\text{e}{-03}$ & $1.779\text{e}{-01}$ & $2.80\text{e}{-03}$ \\
\token{b00bs} & $\leftrightarrow$ & \token{b00b*} & $3.26\text{e}{-02}$ & $4.00\text{e}{-03}$ & $8.77\text{e}{-02}$ & $3.10\text{e}{-03}$ \\
\token{slut} & $\leftrightarrow$ & \token{sh1tter} & $5.62\text{e}{-02}$ & $3.00\text{e}{-03}$ & $1.412\text{e}{-01}$ & $1.80\text{e}{-03}$ \\
\token{faig} & $\leftrightarrow$ & \token{fukker} & $3.13\text{e}{-02}$ & $2.10\text{e}{-03}$ & $6.83\text{e}{-02}$ & $1.00\text{e}{-03}$ \\
\bottomrule
\end{tabular}}

\vspace{0.1em}

\resizebox{\textwidth}{!}{%
\begin{tabular}{>{\raggedleft\arraybackslash}p{0.23\textwidth} @{\hskip 2pt} c @{\hskip 2pt} >{\raggedright\arraybackslash}p{0.23\textwidth} cccc}
\toprule
 &\textbf{Relation} & & \multicolumn{4}{c}{\cellcolor{gray!10}\textbf{Unrelated toxic word pairs}} \\
\midrule
 &  \textbf{Word pair}& & $\ell_2$ (Before) & $\ell_2$ (After) ($\uparrow$) & $\ell_\infty$ (Before) & $\ell_\infty$ (After) ($\uparrow$) \\

\midrule
\token{orgasum} & $\leftrightarrow$ & \token{constructive eviction} & $5.14\text{e}{-02}$ & $2.743\text{e}{-01}$ & $1.086\text{e}{-01}$ & $1.215\text{e}{-01}$ \\
\token{b!tch} & $\leftrightarrow$ & \token{floating bridge} & $7.04\text{e}{-02}$ & $9.92\text{e}{-02}$ & $1.685\text{e}{-01}$ & $5.74\text{e}{-02}$ \\
\token{cock} & $\leftrightarrow$ & \token{biodegradable pollution} & $9.44\text{e}{-02}$ & $5.664\text{e}{-01}$ & $2.417\text{e}{-01}$ & $2.841\text{e}{-01}$ \\
\token{fukker} & $\leftrightarrow$ & \token{karyolysis} & $8.21\text{e}{-02}$ & $4.461\text{e}{-01}$ & $1.928\text{e}{-01}$ & $2.506\text{e}{-01}$ \\
\token{bitches} & $\leftrightarrow$ & \token{fried egg} & $5.57\text{e}{-02}$ & $3.218\text{e}{-01}$ & $1.360\text{e}{-01}$ & $1.412\text{e}{-01}$ \\
\token{penus} & $\leftrightarrow$ & \token{novelette} & $4.47\text{e}{-02}$ & $3.331\text{e}{-01}$ & $9.87\text{e}{-02}$ & $1.576\text{e}{-01}$ \\
\token{kike} & $\leftrightarrow$ & \token{Niobrara River} & $5.22\text{e}{-02}$ & $5.802\text{e}{-01}$ & $1.128\text{e}{-01}$ & $3.029\text{e}{-01}$ \\
\token{b17ch} & $\leftrightarrow$ & \token{dowery} & $5.86\text{e}{-02}$ & $3.491\text{e}{-01}$ & $1.359\text{e}{-01}$ & $1.722\text{e}{-01}$ \\
\token{fukkah} & $\leftrightarrow$ & \token{fish house punch} & $4.83\text{e}{-02}$ & $2.207\text{e}{-01}$ & $1.155\text{e}{-01}$ & $1.266\text{e}{-01}$ \\
\token{basterds} & $\leftrightarrow$ & \token{epenthesis} & $4.85\text{e}{-02}$ & $3.939\text{e}{-01}$ & $1.156\text{e}{-01}$ & $2.164\text{e}{-01}$ \\
\token{fukk} & $\leftrightarrow$ & \token{dry-wood termite} & $7.46\text{e}{-02}$ & $6.182\text{e}{-01}$ & $1.712\text{e}{-01}$ & $2.897\text{e}{-01}$ \\
\token{shyte} & $\leftrightarrow$ & \token{Monrovia} & $5.60\text{e}{-02}$ & $3.725\text{e}{-01}$ & $1.277\text{e}{-01}$ & $2.017\text{e}{-01}$ \\
\token{nazis} & $\leftrightarrow$ & \token{potter around} & $3.59\text{e}{-02}$ & $2.814\text{e}{-01}$ & $8.77\text{e}{-02}$ & $1.552\text{e}{-01}$ \\
\token{cock-sucker} & $\leftrightarrow$ & \token{kind of} & $4.78\text{e}{-02}$ & $2.892\text{e}{-01}$ & $1.116\text{e}{-01}$ & $1.599\text{e}{-01}$ \\
\token{fukkin} & $\leftrightarrow$ & \token{process of monition} & $5.62\text{e}{-02}$ & $3.600\text{e}{-01}$ & $1.249\text{e}{-01}$ & $2.670\text{e}{-01}$ \\
\bottomrule
\end{tabular}}
}

\label{tab:toxicity_related_unrelated}
\end{table*}

\end{document}